\documentclass[11pt]{article}

\usepackage{microtype}
\usepackage{graphicx}
\usepackage{subcaption}
\usepackage{booktabs}
\usepackage{microsoft-tech-report}
\usepackage{hyperref}
\usepackage{amsmath}
\usepackage{amssymb}
\usepackage{mathtools}
\usepackage{amsthm}
\usepackage{bm}
\usepackage{enumitem}
\usepackage[capitalize,noabbrev]{cleveref}
\usepackage{xspace}

\definecolor{color_blue}{HTML}{E7EFFA}
\definecolor{color_green}{HTML}{E6F8E0}
\definecolor{color_gray}{HTML}{ECECEC}
\definecolor{pearDark}{HTML}{2980B9}

\hypersetup{
  colorlinks=true,
  linkcolor=msftblue,
  citecolor=msftblue,
  urlcolor=msftblue,
  pdftitle={World-R1: Reinforcing 3D Constraints for Text-to-Video Generation}
}

\theoremstyle{plain}

\theoremstyle{definition}

\theoremstyle{remark}

\newcommand{\method}{World-R1\xspace}
\techreportshorttitle{World-R1}

\begin{document}
\thispagestyle{empty}

\noindent
\begin{minipage}[c]{0.5\linewidth}
\raggedright
\raisebox{-0.5\height}{\msftbrandmark}
\end{minipage}
\begin{minipage}[c]{0.49\linewidth}
\raggedleft
{\msftdatefont\small\color{msftgray}January 2026}
\end{minipage}\par
\vspace{0.35em}
\noindent{\color{msftline}\rule{\linewidth}{0.8pt}\par}

\vspace{1.0em}
\begin{center}
{{\msfttitlefont\fontsize{21}{25}\selectfont\color{msftdark}
World-R1: Reinforcing 3D Constraints\\
for Text-to-Video Generation\par}}
\vspace{1.25em}

{\normalsize\rmfamily\color{msftdark}
Weijie Wang$^{1,2,*\dagger}$ \hspace{0.9em}
Xiaoxuan He$^{1,*}$ \hspace{0.9em}
Youping Gu$^{1,*}$ \hspace{0.9em}
Yifan Yang$^{2,\ddagger}$\\[-0.1em]
Zeyu Zhang$^{3}$ \hspace{0.9em}
Yefei He$^{1}$ \hspace{0.9em}
Yanbo Ding$^{2}$ \hspace{0.9em}
Xirui Hu$^{3}$\\[-0.1em]
Donny Y. Chen$^{3}$ \hspace{0.9em}
Zhiyuan He$^{2}$ \hspace{0.9em}
Yuqing Yang$^{2,\ddagger}$ \hspace{0.9em}
Bohan Zhuang$^{1,\ddagger}$\par
}
\vspace{0.22cm}

{\footnotesize\rmfamily\color{msftgray}
$^{1}$ Zhejiang University \quad
$^{2}$ Microsoft Research \quad
$^{3}$ Independent Researcher\par
}
\end{center}

\vspace{0.45em}
\begin{msfttitlebox}
\setlength{\parindent}{0cm}
\setlength{\parskip}{0.14cm}
\raggedright
\nohyphens

\begin{abstract}
Recent video foundation models demonstrate impressive visual synthesis but frequently suffer from geometric inconsistencies. While existing methods attempt to inject 3D priors via architectural modifications, they often incur high computational costs and limit scalability. We propose World-R1, a framework that aligns video generation with 3D constraints through reinforcement learning. To facilitate this alignment, we introduce a specialized pure text dataset tailored for world simulation. Utilizing Flow-GRPO, we optimize the model using feedback from pre-trained 3D foundation models and vision-language models to enforce structural coherence without altering the underlying architecture. We further employ a periodic decoupled training strategy to balance rigid geometric consistency with dynamic scene fluidity. Extensive evaluations reveal that our approach significantly enhances 3D consistency while preserving the original visual quality of the foundation model, effectively bridging the gap between video generation and scalable world simulation.
\end{abstract}

\vspace{0.14cm}
{\setlength{\parskip}{0.06cm}\small
{\msftmetalabel{Project Page}\href{https://aka.ms/world-r1}{https://aka.ms/world-r1}\par}
{\msftmetalabel{Correspondence}
\href{mailto:yifanyang@microsoft.com,yuqyang@microsoft.com}{\{yifanyang, yuqyang\}@microsoft.com},
\href{mailto:bohan.zhuang@zju.edu.cn}{bohan.zhuang@zju.edu.cn}\par}
{\msftmetalabel{Conference}The 43$^{rd}$ International Conference on Machine Learning\par}
}
\vspace{0.08cm}
{\footnotesize\rmfamily\itshape\color{msftgray}
$^*$ Equal contribution. \quad
$^{\dagger}$ Work was done during an internship in MSRA. \quad
$^{\ddagger}$ Corresponding authors.\par
}
\end{msfttitlebox}

\begin{figure}[t]
  \centering
  \includegraphics[width=\textwidth]{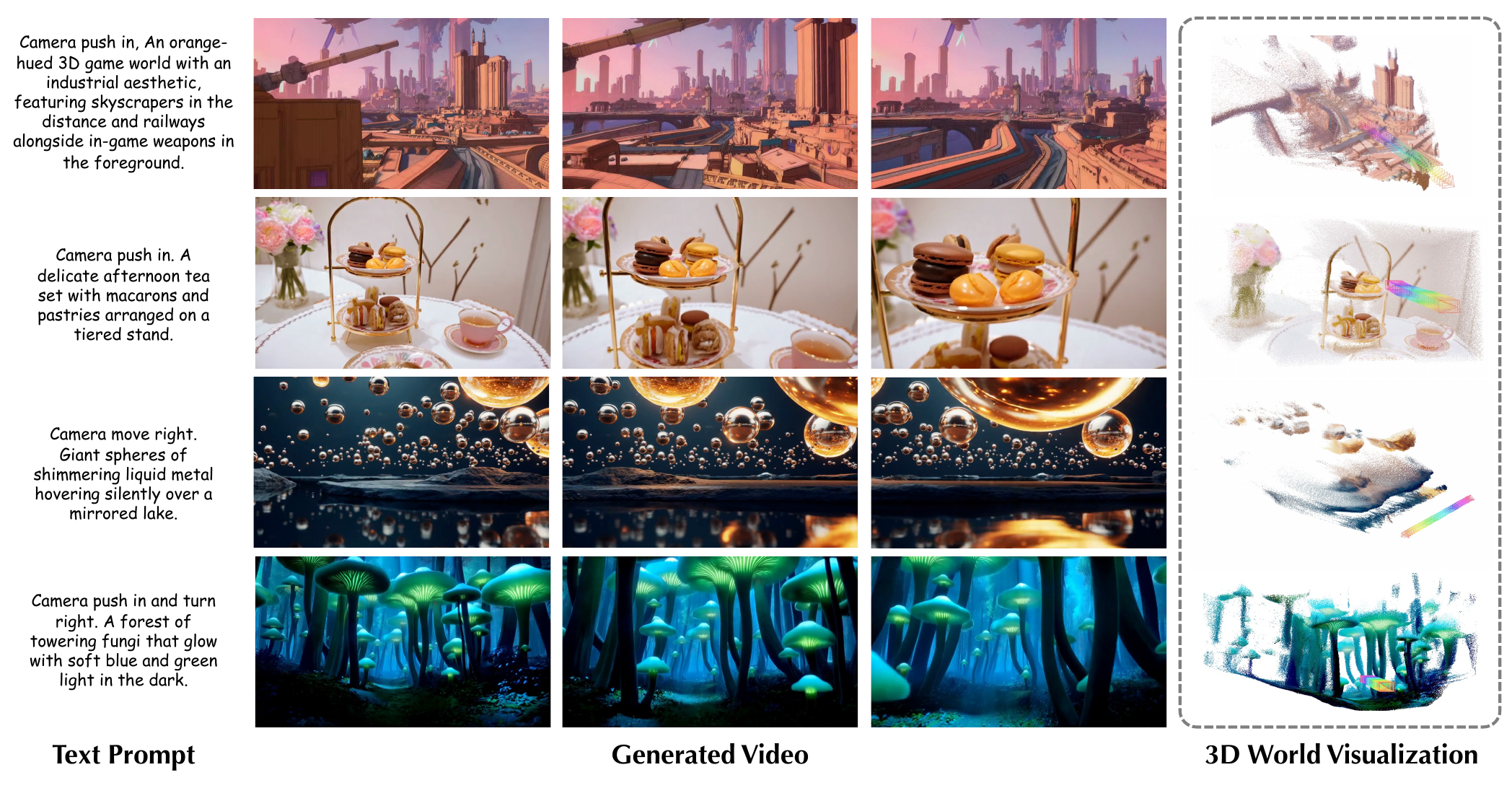}
  \caption{\method enables geometrically consistent world modeling in video foundation models via reinforcement learning. It accepts text input and generates high-quality 3D consistent videos. We visualize the reconstruction results alongside the video frames. Video results are in our supplementary material.}
  \label{teaser}
\end{figure}

\section{Introduction}

Recent advances in visual generation have ignited a paradigm shift, moving beyond simple content creation toward the ambitious goal of world generation. Through extensive pre-training on internet-scale data, video generation models~\cite{yang2025cogvideox,kong2024hunyuanvideo,wan2025wan,blattmann2023stable,chen2024videocrafter2,menapace2024snap} are increasingly recognized as precursors to general-purpose world models. These foundation models demonstrate exceptional capabilities in synthesizing high-fidelity visual environments, holding transformative potential for diverse fields such as autonomous driving, robotics, and immersive content creation.

However, despite their impressive visual proficiency, current video foundation models~\cite{yang2025cogvideox,wan2025wan} fundamentally focus on image-space generation. They lack an intrinsic understanding of the 3D geometry that governs the real world. Consequently, while they excel at generating short, static-camera clips, they frequently suffer from geometric hallucinations and temporal inconsistencies when synthesizing large camera movements or long-horizon driving scenes. Without explicit 3D constraints, objects may morph, vanish, or distort unphysically, revealing that these models are merely mimicking surface-level correlations rather than simulating a coherent real world.

To address these limitations, recent works~\cite{park2025steerx,huang2025voyager,wu2025video,wu2025geometry,song2025worldforge,li2025vmem,chen20254dnex,wu2025genfusion,wang2025drivegen3d} have attempted to bridge video generation with 3D representations. Some approaches~\cite{huang2025voyager,wu2025video,song2025worldforge} explicitly inject 3D priors into the inference process of video models, forcing the generation to adhere to static 3D constraints. While these methods improve consistency to some extent, they often incur prohibitive inference costs and restrict generation precision and generalization scope of the model. Such heavy, inference-time constraints prevent these systems from scaling efficiently or maintaining the fluid diversity of the original video models.

Building upon the finding that video foundation models already inherently encode rich 3D geometric information~\cite{huang2025vidfm3d,an2026vggrpo}, we identify that a path bridging the gap between video generation and world modeling lies in eliciting this latent knowledge rather than merely scaling data or imposing rigid inference constraints. In this paper, we introduce \method, a novel framework that injects world-modeling capabilities into video models via reinforcement learning (RL)~\cite{sutton1998reinforcement}. Uniquely, our approach achieves this without relying on expensive 3D assets for supervised training, and crucially, without altering the model architecture or inference process.

We leverage the inherent consistency requirements of the real world to construct a robust reward mechanism via analysis-by-synthesis. Specifically, we utilize pre-trained 3D foundation models~\cite{wang2026feed,lin2025depth,yang2024depth2,wang2026zpressor,wang2025volsplat,shi2025revisiting} to enforce rigid geometric fidelity through reconstruction and trajectory alignment, while employing vision-language models (VLMs)~\cite{Qwen3-VL,Qwen2.5-VL,Qwen2-VL,Qwen-VL} as semantic critics to evaluate the plausibility of rendered novel views (meta-views). To enable precise motion control without architectural modification, we introduce an implicit camera conditioning strategy that embeds trajectory priors directly into the latent noise. Furthermore, we construct a synthetic pure text dataset to dissociate physical learning from visual bias and adopt a periodic decoupled training strategy, which mitigates the suppression of non-rigid dynamics often caused by strict 3D constraints. Driving this alignment via Flow-GRPO~\cite{liu2025flow}, our framework empowers existing video generation models to internalize geometric laws, effectively transforming them from 2D frame predictors into geometrically consistent world simulators.

To comprehensively evaluate the capabilities of \method, we evaluate both scene generation quality and general video generation performance. Experiments demonstrate that our fine-tuned models significantly improve geometric consistency, achieving an improvement of 10.23dB and 7.91dB on PSNR respectively, while maintaining high scores on general video benchmarks. Additional analyses in the appendix further validate the gains with a reconstruction-independent multi-view consistency metric, dataset-scaling results, long-video evaluation, scene-complexity breakdowns, and comparisons to 3D-aware generation methods. This indicates that our method effectively injects world modeling capability without compromising original generative diversity or visual quality of the model.
Our contributions can be summarized as follows:

\begin{itemize}
    \item We propose \method, a novel paradigm that utilizes RL to align video generation models with 3D foundation models, eliminating the need for cultivating large-scale 3D datasets or computationally expensive inference-time constraints.
    \item We design a comprehensive reward system integrating pre-trained 3D foundation models and VLMs, optimized via Flow-GRPO~\cite{liu2025flow}, allowing the video model to internalize geometric consistency through discriminative feedback.
    \item We constructed a pure text dataset featuring multi-class, multi-level camera control for the post-training of video generation models.
    \item Extensive experiments demonstrate that our framework significantly enhances the world-modeling capabilities of video models, yielding superior 3D reconstruction consistency while preserving high-quality visual generation performance.
\end{itemize}

\section{Related Works}

\noindent\textbf{Controllable Video Generation.}
The emergence of large-scale video foundation models has revolutionized the field of video generation. While early approaches relied on U-Net-based diffusion architectures, recent state-of-the-art models~\cite{kong2024hunyuanvideo,yang2025cogvideox,wan2025wan} have shifted towards DiT architectures. Trained on internet-scale video datasets, these models demonstrate exceptional capabilities in synthesizing high-fidelity visual content. Despite their success, precise camera control within these foundation models remains a persistent challenge. Several methods~\cite{guo2023animatediff,wang2024motionctrl,he2024cameractrl,wu2024motionbooth,kuang2024collaborative,bahmani2025ac3d,zheng2025vidcraft3,bai2025recammaster,li2025realcam,zhang2026panflow} attempt to address this by training auxiliary control modules to introduce explicit camera pose conditioning. However, these approaches primarily focus on trajectory adherence and require additional inputs. Crucially, they fail to guarantee 3D geometric consistency, often resulting in object distortion during complex camera movements. Recognizing that foundation models are inherently insensitive to camera motion without guidance, our approach builds upon the Go-With-The-Flow~\cite{burgert2025go} paradigm. We adopt a module-free strategy for camera control, avoiding the introduction of extra architectural components. Instead, we enhance the instruction following capabilities and 3D consistency through post-training optimization.

\noindent\textbf{3D-Aware Video Generation.}
To mitigate the lack of spatial awareness and 3D consistency in standard video models, recent research has sought to bridge image-to-video(I2V) generation with 3D representations. Several methods~\cite{wu2025video,li2025vmem,huang2025voyager,zhao2025spatia,zhang2025world} attempt to solve this by integrating explicit 3D representations directly into the video generation pipeline. Similarly, Fantasyworld~\cite{dai2025fantasyworld} employs a multi-task learning framework by appending a 3D decoder to the video encoder to generate pointmaps. However, these methods typically necessitate significant architectural modifications and incur high inference latency due to the additional computational burden of 3D modules and can only be applied to I2V tasks. Furthermore, their reliance on static 3D aware datasets~\cite{zhou2018stereo,ling2024dl3dv} for training often severely limits the diversity and dynamic adaptability of the output. In contrast, \method avoids architectural modifications entirely and can handle text-to-video (T2V) tasks. Rather than enforcing 3D constraints via external modules, we leverage reinforcement learning to elicit the latent spatial awareness. This allows the model to learn geometry and maintain consistency without explicit 3D guided inference.

\noindent\textbf{Visual Reinforcement Learning.}
RL~\cite{sutton1998reinforcement} has recently shown immense potential in aligning generative models with human preferences. While PPO~\cite{schulman2017proximal} has been the standard for language models, it incurs prohibitive computational costs when applied to high-dimensional visual data. GRPO~\cite{shao2024deepseekmath} has emerged as a more efficient alternative by eliminating the need for a critic network. Building on this, Flow-GRPO~\cite{liu2025flow} adapts the GRPO framework to flow-matching-based generative models. To further accelerate training, Flow-GRPO-Fast~\cite{liu2025flow} injects noise into the deterministic ODE trajectory at randomly selected intermediate steps, switching to SDE sampling. We extend this framework to the 3D consistent video generation by designing tailored reward mechanisms that specifically penalize geometric inconsistencies. This effectively transforms a general-purpose video generator into a geometrically consistent world simulator.

\section{Preliminaries}

Flow-GRPO~\cite{liu2025flow} enhances flow matching models by integrating online RL to optimize generation quality. While standard flow matching relies on deterministic ODE solvers, RL requires stochasticity for effective exploration and advantage estimation. Flow-GRPO addresses this by reformulating the sampling process and applying GRPO.

\noindent\textbf{Stochastic Sampling via SDE.}
To introduce the necessary stochasticity for RL while preserving the marginal distribution of the pre-trained flow model, Flow-GRPO converts the deterministic flow ODE ($\mathrm{d}\mathbf{x}_t = \mathbf{v}_t \mathrm{d} t$) into a reverse-time Stochastic Differential Equation (SDE):
\begin{equation}
    \mathrm{d} \mathbf{x}_t = \left[\mathbf{v}_t(\mathbf{x}_t) + \frac{\sigma_t^2}{2t}\left(\mathbf{x}_t + (1-t)\mathbf{v}_t(\mathbf{x}_t)\right)\right] \mathrm{d} t + \sigma_t \mathrm{d} \mathbf{w},
    \label{eq:flow_sde}
\end{equation}
where $\sigma_t$ controls the noise level and $\mathbf{w}$ denotes the Wiener process. Discretizing Eq.~\eqref{eq:flow_sde} yields the stochastic update rule that serves as the policy $\pi_\theta$:
\begin{equation}
\label{eq:update_rule}
\mathbf{x}_{t+\Delta t} = \mathbf{x}_t + \bigg[\mathbf{v}_{\theta}(\mathbf{x}_t,t) + \frac{\sigma_t^2}{2t}\big(\mathbf{x}_t + (1-t)\mathbf{v}_{\theta}(\mathbf{x}_t,t)\big)\bigg]\Delta t + \sigma_t\sqrt{\Delta t}\,\epsilon,
\end{equation}
where $\epsilon \sim \mathcal{N}(0,\mathbf{I})$. This formulation transforms the deterministic sampler into a stochastic policy suitable for policy gradient methods.

\noindent\textbf{GRPO.}
Flow-GRPO~\cite{liu2025flow} formulates the denoising process as a Markov Decision Process (MDP). Given a condition $\mathbf{c}$, a group of $G$ trajectories $\{\mathbf{x}^i\}_{i=1}^G$ is sampled. The advantage $\hat{A}^i_t$ for the $i$-th trajectory is estimated by normalizing the reward $R(\mathbf{x}^i_0, \mathbf{c})$ against the group statistics:

\begin{equation}
    \hat{A}_t^i = \frac{R(\bm{x}_0^i, \bm{c}) - \text{mean}(\{R(\bm{x}^i_0,\bm{c})\}_{i=1}^G)}{\text{std}(\{R(\bm{x}_0^i,\bm{c})\}_{i=1}^G)}.
\end{equation}
The model is then updated by maximizing the GRPO objective, which incorporates a KL-divergence constraint to prevent deviation from the reference policy $\pi_{\text{ref}}$:
\begin{equation}
\label{eq:grpo_obj}
\mathcal{J}(\theta) = \mathbb{E}_{\mathbf{c}, \{\mathbf{x}^i\}} \bigg[ \frac{1}{T}\sum_{t=0}^{T-1} \Big( \mathcal{L}_{\text{clip}}(r^i_t, \hat{A}^i_t) - \beta D_{\text{KL}}(\pi_{\theta} || \pi_{\text{ref}}) \Big) \bigg],
\end{equation}
where $\mathcal{L}_{\text{clip}}$ denotes the standard PPO-style clipped surrogate objective, and $r^i_t$ is the probability ratio between the current and old policies. Furthermore, Flow-GRPO employs a denoise reduction strategy, using fewer timesteps during training to accelerate convergence without sacrificing inference quality.

\begin{figure}[t]
  \centering
  \includegraphics[width=\textwidth]{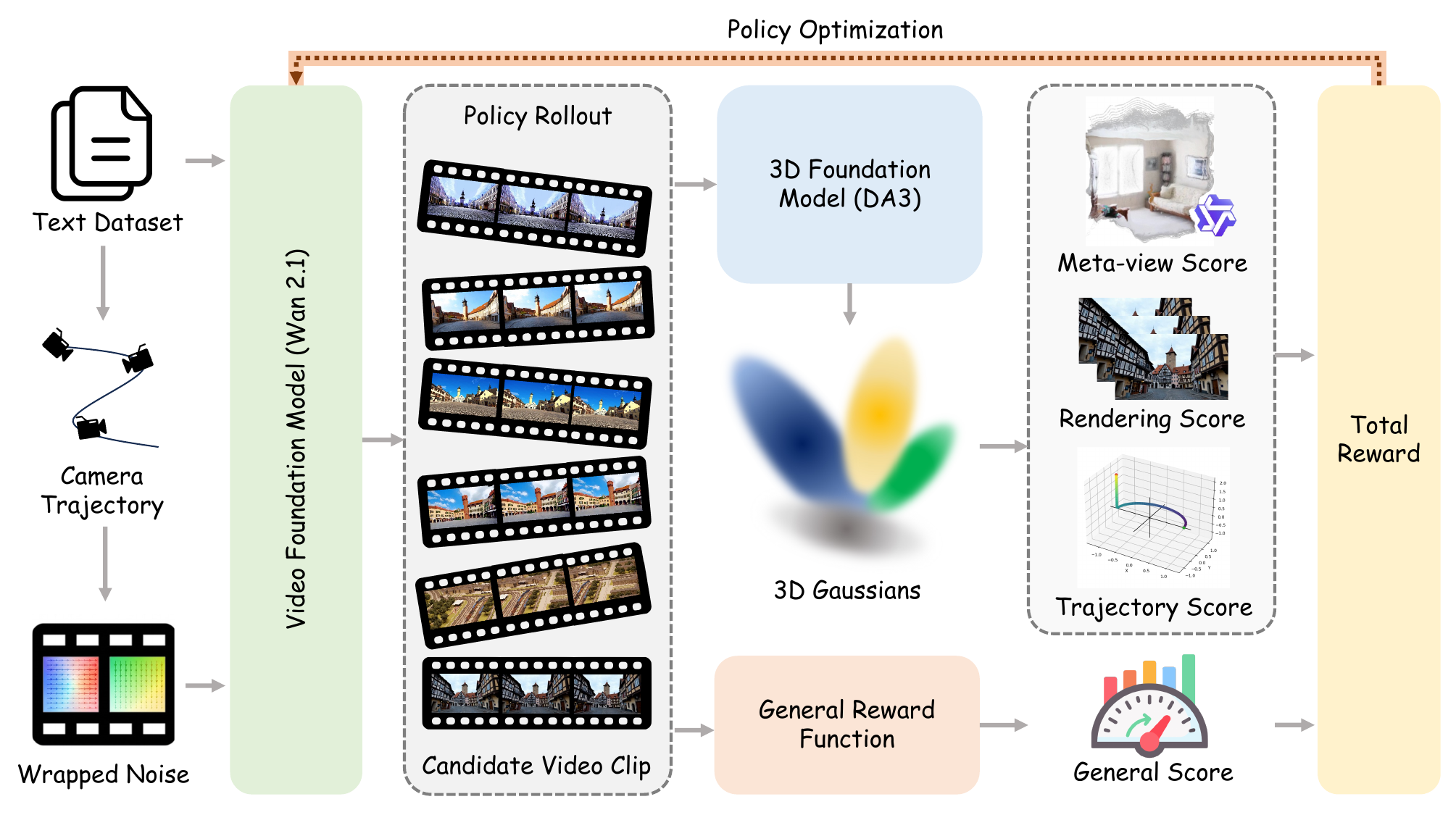}
  \caption{\textbf{Overview of \method.}  \textit{Left:} Given a text prompt, we synthesize a camera trajectory and embed it into the latent space via noise wrapping~\cite{burgert2025go} to achieve implicit camera conditioning. \textit{Middle:} The foundation model~\cite{wan2025wan} generates candidate video clips. \textit{Right:} We employ an comprehensive reward system: the generated video is lifted to a 3D Gaussian Splatting representation via a 3D Foundation Model~\cite{lin2025depth}. We compute a 3D-aware reward score based on meta-view evaluation, rendering fidelity, and trajectory alignment, combined with a general generation reward. The model is optimized using Flow-GRPO-Fast~\cite{liu2025flow} to internalize these geometric constraints.} 
  \label{fig:pipeline}
\end{figure}

\section{Methodology}

\subsection{Overview}

In this section, we detail the proposed framework for injecting world-modeling capabilities into a pre-trained video generation model~\cite{wan2025wan}. The key distinction of our approach lies in its avoidance of explicit architectural modifications and its independence from specialized 3D-aware datasets, instead relying on noise manipulation and reinforcement learning.

\subsection{Camera Conditioning}

Unlike previous methods that require training auxiliary networks~\cite{he2024cameractrl,bai2025recammaster} to encode camera poses, we adopt a parameter-free, implicit conditioning strategy inspired by Go-with-the-Flow~\cite{burgert2025go}. Motivated by the observation that the initial noise distribution in diffusion models significantly influences the trajectory of the generated content, we embed camera motion priors directly into the latent initialization.

\noindent\textbf{Prompt-Driven Trajectory Generation.} We first map the prompt $c$ to a deterministic sequence of camera extrinsic matrices $E = \{E_t\}_{t=0}^N$, where $E_t$ represents the camera pose in 3D space at frame $t$. We define a keyword detection function $\phi(c)$ that scans the input prompt for predefined motion tokens $\mathcal{K} = \{\text{`push in'}, \text{`pan left'}, \text{`orbit left'}, \dots\}$. Based on the detected tokens, we instantiate a trajectory using a parameterized generator. Let $E_0 = I_{4 \times 4}$ be the canonical starting pose. The pose at step $t$ is computed recursively:
\begin{equation}
E_t = E_{t-1} \cdot T_{\text{action}}(t),
\end{equation}
where $T_{\text{action}}$ denotes the transformation matrix specific to the motion type. 
When multiple camera movement instructions are present in the prompt, we concatenate the generated trajectories to obtain the final trajectory.

\noindent\textbf{Trajectory-to-Flow Projection.} To warp the noise representations, we project the 3D camera trajectory $E$ into a sequence of 2D dense optical flow fields. We adopt a pinhole camera model and approximate the scene geometry as a fronto-parallel plane located at a constant reference depth $z_{\text{ref}}$. For a pixel $u \in \mathbb{R}^2$ in frame $t$, its projected coordinate $u'$ in frame $t+1$ is derived via the planar homography induced by the relative camera motion:
\begin{equation}
u' \sim K \left( R_{\text{rel}} + \frac{1}{z_{\text{ref}}} \mathbf{t}_{\text{rel}} \mathbf{n}^\top \right) K^{-1} u,
\end{equation}
where $K$ is the intrinsic matrix, $(R_{\text{rel}}, \mathbf{t}_{\text{rel}})$ denotes the relative rigid body transformation $E_{t+1} E_t^{-1}$, and $\mathbf{n} = [0, 0, 1]^\top$ is the normal vector of the image plane. This process yields the forward optical flow field defined as $f(u) = u' - u$.

\noindent\textbf{Discrete Noise Transport.} While the projected flow field $f(u)$ defines a continuous motion prior, the initial noise is defined on a discrete spatial grid. Directly warping noise using $f(u)$ would lead to variance collapse in overlapping regions and missing noise in disoccluded areas, violating the standard Normal distribution required by diffusion-based models. To address this, we adopt the discrete noise transport mechanism from Go-with-the-Flow~\cite{burgert2025go}, which formulates noise warping as a mass transport problem on a bipartite graph induced by the flow field.

Specifically, the continuous flow $f(u)$ induces discrete correspondences $v \rightarrow v'$ between source pixels $v$ at frame $t$ and target pixels $v'$ at frame $t+1$. Noise values are aggregated according to these correspondences, while a density tracker $\rho(v')$ records the number of incoming contributions. The transported noise is then normalized to preserve unit variance:
\begin{equation}
z_{t+1}(v') = \frac{1}{\sqrt{\rho(v')}} \sum_{v \rightarrow v'} z_t(v),
\end{equation}
where $z_{t+1}$ is the noise value at a target pixel $v'$. This discrete transport scheme ensures that camera-induced spatial structure is injected into the initial noise while maintaining its standard Normal distribution across frames.

\subsection{Reward Design}

To ensure the generated videos enhance 3D capabilities while maintaining high visual fidelity, we design a composite reward mechanism. This objective function $R$ is formulated as a weighted aggregation of a physics-grounded 3D consistency term $R_{\text{3D}}$ and a general quality assessment term $R_{\text{gen}}$:
\begin{equation}
R(\mathbf{x}, c) = R_{\text{3D}}(\mathbf{x}, E, c) + \lambda_{\text{gen}} R_{\text{gen}}(\mathbf{x}, c).
\label{eq:total_reward}
\end{equation}
Here $\mathbf{x}$ represents the generated video, $E$ denotes the target camera trajectory derived from the prompt condition $\mathbf{c}$, and $\lambda_{\text{gen}}$ is a balancing hyper-parameter.

\noindent\textbf{3D-Aware Reward.} We employ an analysis-by-synthesis strategy to distinguish between genuine 3D parallax and 2D content drift. We leverage the Depth Anything 3~\cite{lin2025depth} which directly reconstructs the scene geometry as a 3D Gaussian Splatting (3DGS) representation $\Phi_{\text{GS}}$ and estimates the corresponding camera trajectory $\hat{E}$ from the generated video $\mathbf{x}$. We define the 3D-aware reward $R_{\text{3D}}$ as a linear combination of three distinct consistency metrics tailored to geometry, appearance, and control accuracy:
\begin{equation}
R_{\text{3D}} = \mathcal{S}_{\text{meta}} + \mathcal{S}_{\text{recon}} + \mathcal{S}_{\text{traj}}.
\label{eq:3d_reward}
\end{equation}
Specifically, the geometric integrity term $\mathcal{S}_{\text{meta}}$ is computed by rendering 3D Gaussians from a novel meta-view. The renderings are then evaluated by Qwen3-VL~\cite{Qwen3-VL} to assess text fidelity and structural reliability, effectively penalizing artifacts that remain occluded in the canonical view but indicate underlying geometric flaws. To ensure visual consistency, $\mathcal{S}_{\text{recon}}$ measures pixel-level fidelity by comparing the generated video $\mathbf{x}$ against its re-rendered counterpart from the 3DGS representation $\Phi_{\text{GS}}$, quantifying the similarity via the negated perceptual distance ($1 - \text{LPIPS}$~\cite{zhang2018unreasonable}). Finally, the trajectory alignment term $\mathcal{S}_{\text{traj}}$ assesses control precision by calculating the deviation between the generated camera condition $E$ and the predicted trajectory $\hat{E}$, thereby ensuring the generated camera motion adheres strictly to the user instructions. For more details of the reward, please refer to \cref{sec:reward_details}.

\noindent\textbf{General Generation Reward.} We further incorporate a general reward function $R_{\text{gen}}$ to guarantee the visual quality and aesthetic appeal of the generated content. We formulate this objective as the average aesthetic assessment across the first $K$ frames:
\begin{equation}
R_{\text{gen}}(\mathbf{x}) = \frac{1}{K} \sum_{t=0}^{K-1} \mathcal{H}(\mathbf{x}_t),
\end{equation}
\noindent where $\mathcal{H}$ denotes the HPSv3~\cite{ma2025hpsv3} score evaluated on each frame $\mathbf{x}_t$. By maximizing this aggregated score, we ensure the video consistently aligns with human visual preferences and maintains high perceptual standards throughout the sequence.

\subsection{Dataset Preparation}
\label{sec:dataset}

A significant bottleneck in previous camera-control research has been the reliance on open-domain video datasets, where limited resolution and noisy text-video alignment often hinder precise control learning. To overcome this limitation and dissociate the learning of 3D constraints from specific video distributions, we construct a \textit{Pure Text Dataset} specifically tailored for world simulation.

We leverage the advanced instruction-following and creative capabilities of Gemini~\cite{team2023gemini,team2024gemini} to synthesize a diverse corpus of high-quality scene descriptions. The core advantage of this dataset lies in its independence from fixed visual priors, allowing the model to synthesize a vast array of distinct scenarios, ranging from natural landscapes and urban structures to surrealist environments. 
The dataset comprises approximately 3,000 unique entries, systematically categorized to cover a wide spectrum of visual domains and physical properties. 
Furthermore, we categorize these prompts by control complexity, including samples with implicit motion, single directional commands, and complex composite trajectories. This multi-level structure enables the model to learn physics-compliant generation across varying degrees of difficulty. For more details about the dataset, please refer to \cref{sec:data_details}.

\subsection{Training Strategy}
\label{sec:train}

Strict adherence to 3D consistency can inadvertently suppress the generation of non-rigid dynamics, such as moving characters or naturally deforming objects. To balance geometric fidelity with the capability for dynamic scene generation, we adopt a periodic decoupled training strategy.

Our training datasets contain approximately 500 prompts specifically describing highly dynamic scenes. During training, we implement a multi-stage cycle. In the primary stage, the model is trained with the full weighted reward to enforce 3D-aware capability. However, to prevent the model from overfitting to static geometric constraints, we introduce a dynamic fine-tuning phase after every 100 training steps. In this phase, we temporarily disable the 3D-aware reward $R_{\text{3D}}$ and optimize the model exclusively on the dynamic data subset using only the general reward $R_{\text{gen}}$. This mechanism acts as a regularizer, ensuring the model retains its generalization power for complex dynamic motions while learning world simulation. The efficacy of this strategy is verified in \cref{sec:ablation}. Visualization results are illustrated in \cref{sec:more_vis}.

\section{Experiments}
\label{sec:exp}

\begin{figure}[t]
  \centering
  \includegraphics[width=\textwidth]{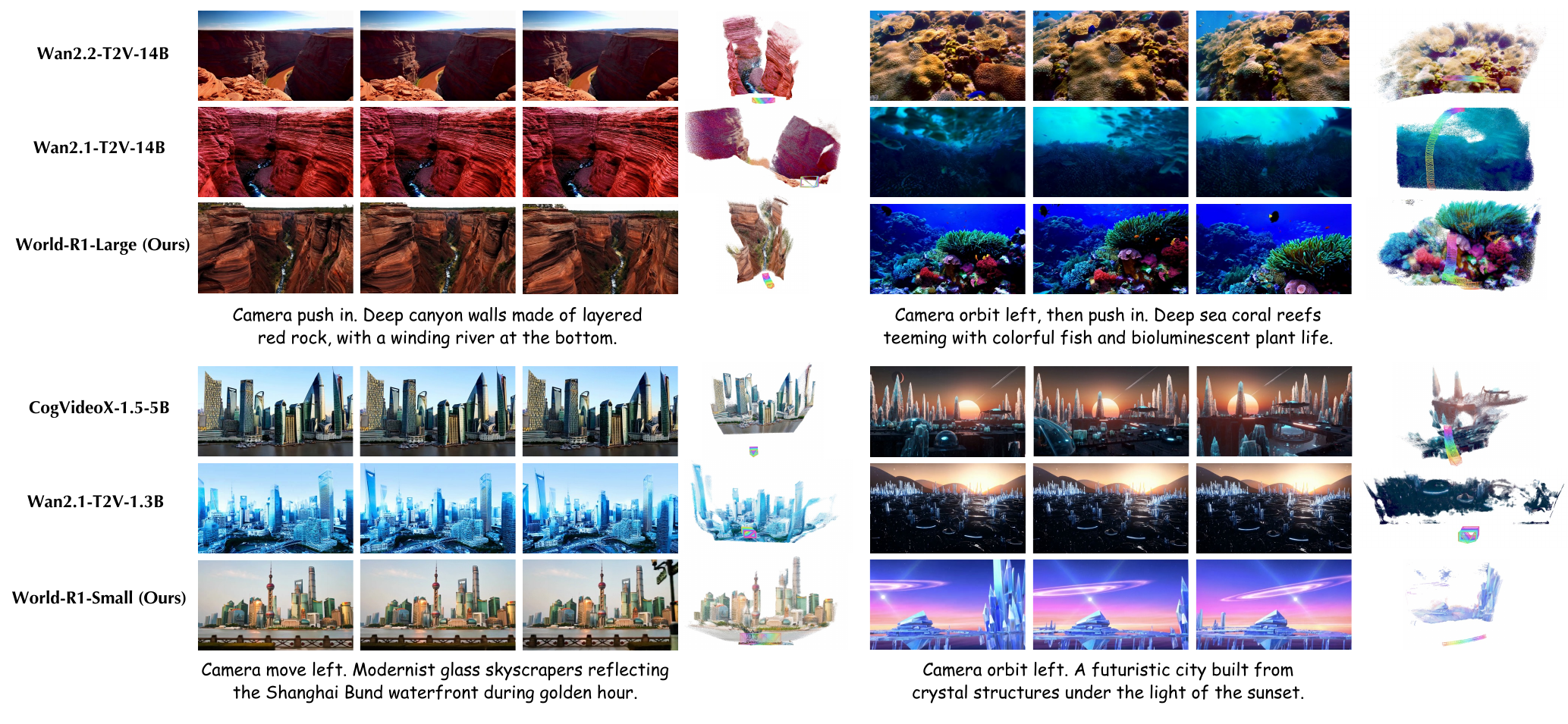}
  \caption{\textbf{Qualitative Comparison.} We visualize the generated video frames and the corresponding 3D reconstruction for diverse scenes involving different camera motions. Compared to the video foundation models~\cite{wan2025wan,yang2025cogvideox}, \method demonstrates superior 3D consistency. Our model preserves strict scene permanence and geometry performance, yielding dense and structured 3D visualizations that confirm its robust world-modeling capabilities.} 
  \label{fig:visualization} 
\end{figure}

\subsection{Experimental Settings}

\noindent\textbf{Implementation Details.}
We leverage the state-of-the-art open-source video generation models Wan 2.1 1.3B and Wan 2.1 14B~\cite{wan2025wan} as backbone foundations. By fine-tuning on our proprietary dataset (\cref{sec:dataset}), we obtain two model variants.
\textit{\method-Small} is trained using 48 NVIDIA H200 GPUs, while \textit{\method-Large} is trained using 96 NVIDIA H200 GPUs. Both models are optimized at a video resolution of $832 \times 480$. For reinforcement learning, we adopt Flow-GRPO-Fast~\cite{liu2025flow} across both experimental settings, utilizing 48 parallel groups with a group size of 8 to ensure training stability and scalability. For more details, please refer to \cref{sec:train_details}.

\noindent\textbf{Evaluation Metrics.}
We employ a dual-pronged evaluation protocol focusing on both geometric integrity and general generative quality to assess world-modeling capabilities. We first introduce a reconstruction-based metric rooted in 3D geometry to quantify 3D consistency. We employ 3DGS~\cite{kerbl20233d,ye2025gsplat} to reconstruct the underlying scene geometry for each generated video. We then re-render the scene from the exact camera trajectory specified in the input prompt. We measure the extent to which the video maintains structural coherence by calculating the PSNR, SSIM~\cite{wang2004image}, and LPIPS~\cite{zhang2018unreasonable} between the generated video and its 3D-consistent re-rendering. High reconstruction fidelity indicates that the model has successfully suppressed geometric hallucinations and temporal warping. To reduce dependence on the reconstruction pipeline, we additionally report a reconstruction-independent Multi-View Consistency Score (MVCS) in \cref{sec:additional_exp}. Furthermore, we utilized VBench~\cite{huang2024vbench} as a standardized benchmark to evaluate general video quality. We report a comprehensive suite of sub-metrics including Aesthetic Quality, Imaging Quality, Motion Smoothness, and Consistency scores. This ensures that our alignment process improves the visual fidelity and dynamic range of the base foundation model.

\subsection{SoTA Comparisons and Analysis}

\begin{table}[t]
\centering
\caption{Evaluation of general video generation quality on VBench~\cite{huang2024vbench}. Due to resource constraints, we did not test the performance of World-R1-Large. Best scores are in \colorbox{pearDark!20}{blue}, second-best in \colorbox{color_green}{green}, third-best in \colorbox{yellow!20}{yellow}.}
\label{tab:vbench_results}
\resizebox{\textwidth}{!}{%
\begin{tabular}{l|ccccc}
\toprule
\textbf{Method}
& \shortstack{\textbf{Aesthetic}\\\textbf{Quality} $\uparrow$}
& \shortstack{\textbf{Imaging}\\\textbf{Quality} $\uparrow$}
& \shortstack{\textbf{Motion}\\\textbf{Smoothness} $\uparrow$}
& \shortstack{\textbf{Subject}\\\textbf{Consistency} $\uparrow$}
& \shortstack{\textbf{Background}\\\textbf{Consistency} $\uparrow$} \\
\midrule
CogVideoX-1.5-5B~\cite{yang2025cogvideox}
& \colorbox{yellow!20}{62.07} & \colorbox{yellow!20}{65.34} & 98.15 & \colorbox{color_green}{96.56} & \colorbox{color_green}{96.81} \\

Wan2.1-T2V-1.3B~\cite{wan2025wan}
& \colorbox{color_green}{62.43} & \colorbox{color_green}{66.51} & 97.44 & \colorbox{yellow!20}{96.34} & \colorbox{pearDark!20}{97.29} \\

\midrule
GCD~\cite{van2024generative}
& 38.21 & 41.56 & 98.37 & 88.94 & 92.00 \\

Trajectory-Attention~\cite{xiao2024trajectory}
& 38.50 & 51.00 & 98.21 & 90.60 & 92.83 \\

DAS~\cite{gu2025diffusion}
& 39.86 & 51.55 & \colorbox{color_green}{99.14} & 90.34 & 92.03 \\

ReCamMaster~\cite{bai2025recammaster}
& 42.70 & 53.97 & \colorbox{pearDark!20}{99.28} & 92.05 & 93.83 \\
\midrule
\textbf{\method-Small (Ours)}
& \colorbox{pearDark!20}{65.74} & \colorbox{pearDark!20}{67.53} & \colorbox{yellow!20}{98.55} & \colorbox{pearDark!20}{97.58} & \colorbox{yellow!20}{96.67} \\

\bottomrule
\end{tabular}
}
\end{table}

\begin{table}[t]
\centering
\captionsetup{width=\textwidth}
\caption{\textbf{Quantitative comparison of 3D consistency.} We evaluate the structural stability of generated videos by reconstructing the scene geometry via 3DGS~\cite{kerbl20233d} and re-rendering it on the test set proposed in \cref{sec:dataset}. Notably, \method significantly outperforms baseline foundation models, 
achieving
a substantial reduction in geometric hallucinations and improved 3D consistency.}
\label{tab:3d_results}
\resizebox{0.62\textwidth}{!}{%
\begin{tabular}{l|ccc}
\toprule
\textbf{Method} & \textbf{PSNR} $\uparrow$ & \textbf{SSIM} $\uparrow$ & \textbf{LPIPS} $\downarrow$ \\
\midrule
CogVideoX-1.5-5B~\cite{yang2025cogvideox} 
& 24.44 & 0.783 & 0.242 \\

Wan2.2-T2V-14B~\cite{wan2025wan} 
& 23.47 & 0.779 & 0.253 \\

Wan2.2-T2V-5B~\cite{wan2025wan} 
& 22.36 & 0.716 & 0.303 \\

Wan2.1-T2V-14B~\cite{wan2025wan} 
& 19.76 & 0.629 & 0.405 \\

Wan2.1-T2V-1.3B~\cite{wan2025wan} 
& 17.40 & 0.550 & 0.467 \\
\midrule
\textbf{\method-Small (Ours)} 
& 27.63 & 0.858 & 0.201 \\
\textbf{\method-Large (Ours)} 
& \textbf{27.67} & \textbf{0.865} & \textbf{0.162} \\
\bottomrule
\end{tabular}
}
\end{table}

We benchmark \method against leading foundational video generation models including Wan 2.1~\cite{wan2025wan} and CogVideoX~\cite{yang2025cogvideox}. We also compare our approach with state-of-the-art auxiliary control methods such as CameraCtrl~\cite{he2024cameractrl} and ReCamMaster~\cite{bai2025recammaster} to contextualize our contributions further. These methods introduce explicit architectural modules for camera guidance.

\noindent\textbf{Quantitative Analysis.}
\cref{tab:3d_results} presents the performance of \method on 3D consistency. Our approach significantly outperforms all baselines and achieves a PSNR improvement of \textbf{10.23dB} and \textbf{7.91dB} along with a substantial improvement in SSIM and LPIPS. This quantitative leap confirms that our RL-based alignment successfully injects rigid geometric constraints into the video generation process and mitigates the artifacts often observed in standard foundation models. Auxiliary control methods such as CameraCtrl~\cite{he2024cameractrl} often suffer from texture distortion and lead to lower reconstruction scores despite adhering to trajectories.

\cref{tab:vbench_results} details the performance on VBench~\cite{huang2024vbench}. \method surpasses the performance of the original Wan 2.1~\cite{wan2025wan} backbone. Meanwhile, compared to models controlled by explicit cameras, we far surpass them in Aesthetic Quality, Imaging Quality, and Subject Consistency. \cref{sec:additional_exp} provides additional robustness analyses, including dataset scaling, 121-frame evaluation, category-wise results for complex scenes, and a consolidated comparison with 3D-conditioned and camera-control methods. These results demonstrate that our framework effectively prevents the degradation of visual quality often associated with heavy structural constraints.

\begin{figure}[t]
  \centering
  \includegraphics[width=\textwidth]{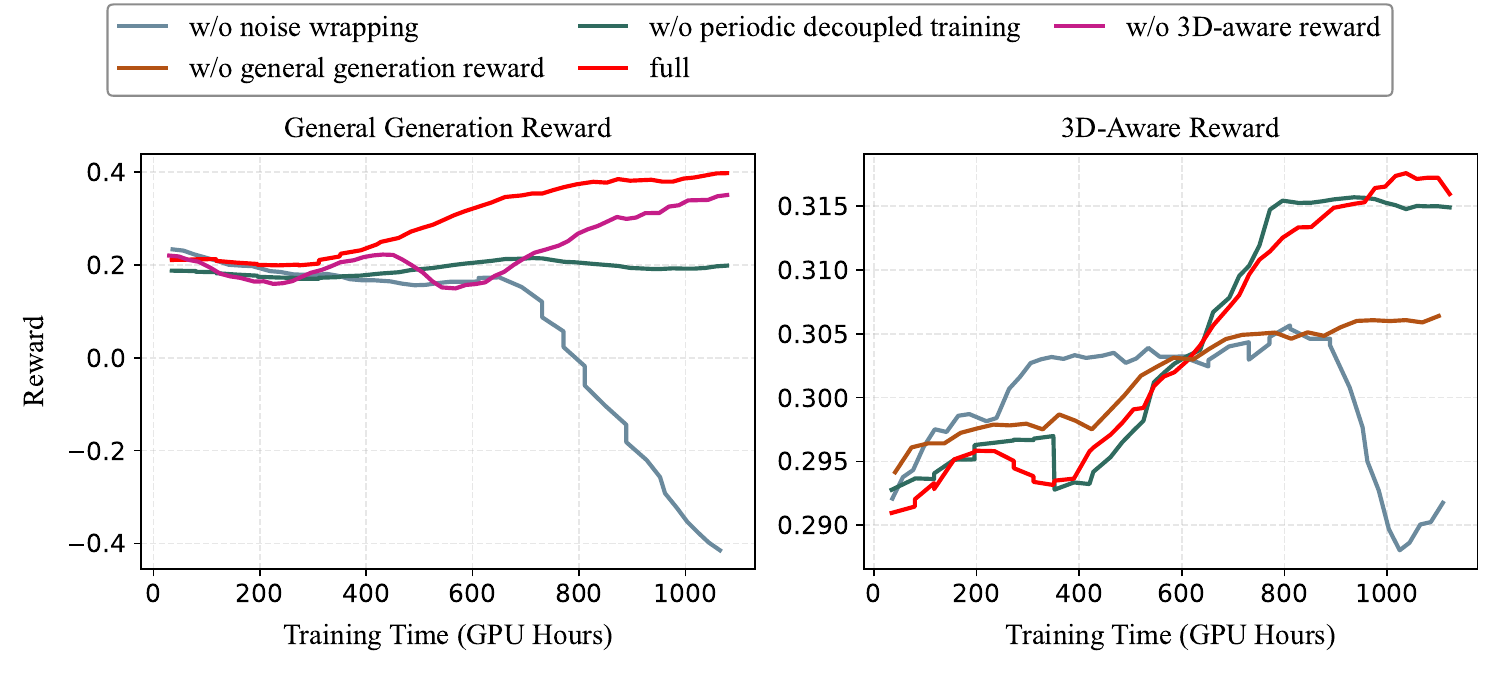}
  \caption{\textbf{Ablation study.} We visualize the evolution of the general generation reward and the 3D-aware reward during fine-tuning of World-R1-Small. The curves demonstrate the contribution of each component.} 
  \label{fig:ablation} 
\end{figure}

\noindent\textbf{Qualitative Analysis.}
Visual comparisons presented in \cref{fig:visualization} illustrate the capability gap. Baseline models frequently exhibit unrealistic transitions where objects vanish or walls warp in scenarios requiring complex camera motions such as circling a building or pushing through a corridor. \method maintains strict object permanence and rigid geometry in contrast. The 3DGS reconstruction visualizations further substantiate this finding. The point clouds derived from our videos are dense and structured, while reconstructions from baseline videos appear sparse and noisy due to multi-view inconsistencies.

\noindent\textbf{User Study.}
To evaluate human subjective preference, we conducted a blind user study comparing World-R1 against the base model Wan 2.1~\cite{wan2025wan} (specifically, Wan2.1-T2V-1.3B vs. World-R1-Small and Wan2.1-T2V-14B vs. World-R1-Large). We curated a test set of 30 complex prompts covering various domains (e.g., architectural fly-throughs, natural landscapes) and camera motions (e.g., push in, orbit). 25 participants were presented with randomized side-by-side video pairs and asked to select the superior model based on three criteria: (1) Geometric Consistency (stability of objects and structures), (2) Camera Control Accuracy (adherence to the specified trajectory), and (3) Overall Visual Quality.
\method demonstrates a commanding lead in both geometric consistency and camera control accuracy, achieving win rates of 92\% and 76\% respectively over Wan 2.1~\cite{wan2025wan}. Notably, in terms of overall preference, \method was chosen as the best video in 86\% of comparisons, validating that our RL-based alignment effectively bridges the gap between creative generation and geometric reasonableness. 

\subsection{Ablation Study}
\label{sec:ablation}

We perform ablation studies on \method-Small to verify the effectiveness of our key components as shown in \cref{fig:ablation}, systematically isolating the contributions of our reward functions, model conditioning, and data training strategy. For \textbf{reward mechanism}, our analysis confirms that the 3D-aware reward $R_{3D}$ is fundamental for establishing geometric consistency, whereas the general generation reward $R_{gen}$ proves indispensable for maintaining high perceptual fidelity and preventing the aesthetic degradation. For \textbf{model conditioning}, we observed that removing the implicit camera injection via noise wrapping leads to significantly slower convergence and inferior trajectory alignment, demonstrating that embedding motion priors directly into the latent initialization provides a critical inductive bias for the optimization process. For \textbf{training strategy}, without the intermittent relaxation of constraints using the dynamic data subset, the model overfits to static rigidity and suppresses natural non-rigid dynamics, thereby validating our approach of using dynamic fine-tuning as a regularizer to balance strict world simulation with fluid motion generation. \cref{sec:reward_hacking} further analyzes reward hacking and reports component-wise reward ablations.

\section{Conclusion}
We presented \method, a scalable paradigm for endowing video generation models with robust world-modeling capabilities. By reformulating the alignment of video generation and 3D geometry as a reinforcement learning problem, we successfully elicit latent spatial awareness from pre-trained models without requiring explicit 3D architectural modules or expensive supervised datasets. Our approach leverages a composite reward system grounded in multi-view consistency and semantic coherence, optimized via Flow-GRPO, to ensure physical validity while maintaining visual fidelity. The proposed implicit camera conditioning and periodic training strategy further enable precise trajectory control without compromising the generation of dynamic non-rigid content. Quantitative and qualitative evaluations confirm that our method significantly improves geometric consistency and preserves general video quality compared to state-of-the-art baselines. By effectively transforming video generators into geometrically consistent simulators, our framework opens new avenues for applications in autonomous driving simulation and physical world modeling.

\noindent\textbf{Limitations and Future Work.} 
While World-R1 successfully injects world-modeling capabilities into video foundation models, two limitations remain. First, the computational cost of applying reinforcement learning to video generation is still a significant bottleneck. Unlike supervised fine-tuning, online RL requires repeated video rollouts and reward evaluation, making the training process more expensive than standard post-training pipelines. Developing more efficient rollout strategies, lower-cost reward evaluation, and stable video RL optimization will therefore be important directions for future work. Second, World-R1 is built on top of existing video foundation models and is consequently bounded by their generative capacity. Challenging cases such as dense multi-object composition, fine-grained non-rigid motion, detailed hand dynamics, and very long-horizon scene evolution may still inherit artifacts from the base model. As stronger video foundation models become available, our post-training framework can directly benefit from their improved scene understanding and motion generation capabilities.

\section*{Impact Statement}

This paper introduces a method to improve the geometric consistency of video generation models by aligning video synthesis with 3D constraints through reinforcement learning and flow-matching optimization. This advancement enhances the reliability of generated videos for applications requiring high physical accuracy, such as physical world modeling and autonomous driving simulation.
The societal and ethical considerations of this work align with those of existing high-fidelity video generation techniques. The approach does not introduce novel model architectures or data modalities that would create new ethical risks beyond known issues like the potential misuse of synthetic content. By utilizing a fast training strategy and a text-based dataset, the method offers a computationally efficient alternative, potentially reducing the carbon footprint associated with large-scale model training. No additional broader impacts requiring specific discussion are identified.

\section*{Author Contributions}

Weijie Wang led the overall project, including the end-to-end framework design, code implementation, paper writing, and experimental validation. Xiaoxuan He was responsible for large-scale experiment debugging, validation, and experimental optimization. Youping Gu contributed to the development and debugging of the camera conditioning module and conducted part of the testing experiments. Zeyu Zhang, Yefei He, Yanbo Ding and Xirui Hu contributed to experimental analysis, evaluation design and manuscript refinement. Yifan Yang, Donny Y. Chen, Zhiyuan He, Yuqing Yang and Bohan Zhuang advised the project, provided high-level research guidance, supervised the study, and contributed to the final manuscript revision.

\bibliography{reference}
\bibliographystyle{unsrtnat}

\newpage
\appendix
\setcounter{table}{0}
\setcounter{figure}{0}
\setcounter{section}{0}
\renewcommand{\thetable}{\Alph{table}}
\renewcommand{\thefigure}{\Alph{figure}}
\renewcommand{\thesection}{\Alph{section}} 
\section{Implementation Details}

\subsection{Reward Formulation and Details}
\label{sec:reward_details}

The core of our alignment strategy is the 3D-aware reward, which utilizes an analysis-by-synthesis approach to penalize geometric inconsistencies. Given a generated video clip $x$, we employ Depth Anything 3 to lift the sequence into a 3D Gaussian Splatting (3DGS) representation, denoted as $\Phi_{GS}$, and simultaneously estimate the camera trajectory $\hat{E}$. The reward $R_{3D}$ is computed as the linear combination of three distinct metrics as in \cref{eq:3d_reward}: $R_{3D} = \mathcal{S}_{meta} + \mathcal{S}_{recon} + \mathcal{S}_{traj}$, all rewards are limited to $[0,1]$ and added directly. 
For the final score calculation, the value range of $R_{gen}$ is $[-1,1]$, and the value range of $R_{3D}$ is $[0,3]$. We will add them directly with $\lambda_{\text{gen}}=1$ according to \cref{eq:total_reward}.

The \textbf{Geometric Integrity Score ($\mathcal{S}_{meta}$)} is the most critical component for detecting hallucinations that appear plausible in 2D but fail in 3D (e.g., flat ``billboard" objects). We render the optimized $\Phi_{GS}$ from a novel \textit{meta-view}. a camera pose significantly offset from the generation trajectory (e.g., backward from the origin). From this perspective, we can have a more comprehensive observation of the whole scene. We employ Qwen3-VL~\cite{Qwen3-VL} as a semantic critic to evaluate the structural plausibility of this meta-view. The VLM is prompted to act as a graphics expert, identifying artifacts such as floaters, geometric distortion, or texture stretching. The specific system instruction provided to the VLM is detailed below:

\begin{center}
\fbox{\begin{minipage}{0.9\textwidth}
\noindent\textbf{System Instruction:} You are a professional 3D vision expert. I used a text prompt to generate a video and reconstructed a corresponding 3D Pointmap from the video.

\noindent\textbf{Original Prompt:} {[Text Prompt]}

\noindent\textbf{Criteria:} 
Your task is to judge the quality of the original video by analyzing the provided image of its resulting 3D pointmap. A good video (smooth, orbiting camera) creates a good pointmap. A bad video (static, jittery, or zooming) creates a bad pointmap.
Please provide a score from 0 to 9 based on these criteria:

\begin{itemize}[itemsep=0pt, topsep=0pt]
    \item 9: Excellent - A dense, clean, and complete 3D model. Perfect 360° orbital motion, high stability.
    \item 7-8: Good - A clear object with strong 3D structure. May have minor holes or noise. Good, smooth camera arc with strong parallax.
    \item 4-6: Mediocre - Object is recognizable, but the map is sparse, noisy, or "flat" (lacks 3D depth). Poor parallax (e.g., just a zoom or pan instead of an orbit), or the video was jittery, blurry, or had object/lighting inconsistencies.
    \item 2-3: Poor - A chaotic jumble of points or a simple 2D projection. Static camera (no motion) or completely unstable.
    \item 0-1: Very Poor - Empty or just random noise. Unusable.
\end{itemize}

\noindent\textbf{Scoring:} 
Output only a single digit (0-9):
\end{minipage}}
\end{center}

\begin{figure}[t]
  \centering
  \includegraphics[width=\textwidth]{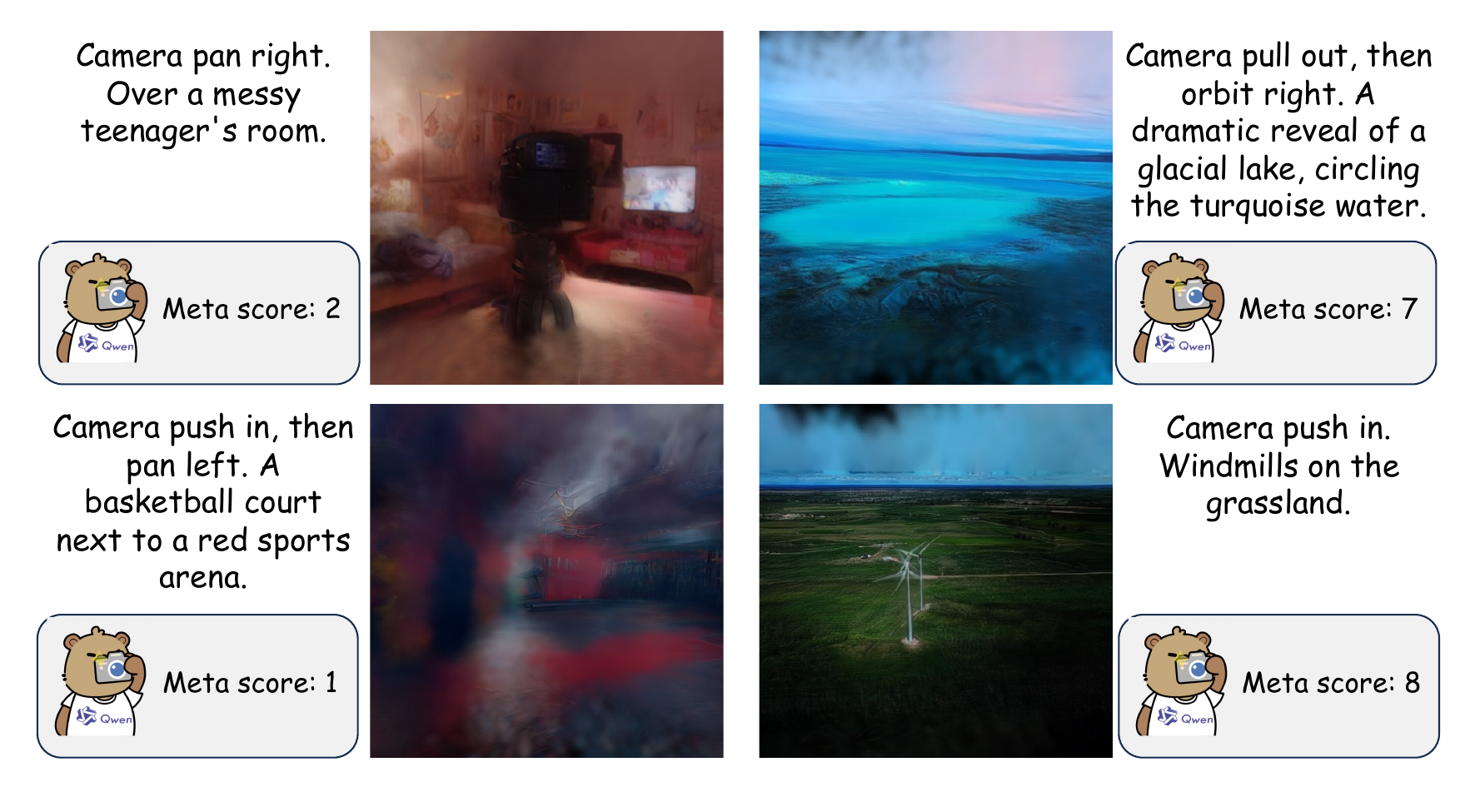}
  \caption{\textbf{Meta view visualization.} Left: Due to unstable camera trajectory and poor 3D consistency, the reconstructed 3DGS cannot be displayed stably. On the right: A high-quality video with a good meta view and consistent with the prompt description.} 
  \label{fig:appendix_meta} 
\end{figure}

To ensure the meta-view evaluation is robust, we visualize the difference between the good meta-view and bad meta-view in \cref{fig:appendix_meta}. The canonical view usually appears structurally sound, whereas the meta-view reveals the "cardboard prop" effect in baseline models, which our reward explicitly suppresses. We multiply the obtained score by 0.1 to obtain the final score, ensuring the score falls within the range $[0,1]$.

The \textbf{Reconstruction Fidelity Score ($\mathcal{S}_{recon}$)} ensures that the lifted 3D representation remains faithful to the original generated content. We re-render the scene from the estimated trajectory $\hat{E}$ to obtain the reconstruction $\hat{x}$ and measure the pixel-level similarity using LPIPS. This term is calculated as $\mathcal{S}_{recon} = 1 - \text{LPIPS}(x, \hat{x})$, favoring generations that allow for photorealistic 3D reconstruction.

The \textbf{Trajectory Alignment Score ($\mathcal{S}_{traj}$)} quantifies the controllability of the generation. It measures the deviation between the user-specified input camera trajectory $E$ (embedded via our implicit conditioning) and the actual camera motion $\hat{E}$ derived from the video. We compute this as the negative exponential of the $L_2$ distance for translation and the geodesic distance for rotation, ensuring the generated camera movement strictly adheres to the prompt instructions.

\subsection{More Training Details}
\label{sec:train_details}

\noindent\textbf{Infrastructure and Base Models.} Our experimental framework is built upon the state-of-the-art Wan 2.1 video foundation models \citep{wan2025wan}. We train two variants of our model: \textit{World-R1-Small}, initialized from Wan2.1-T2V-1.3B, and \textit{World-R1-Large}, initialized from Wan2.1-T2V-14B. The training infrastructure leverages a high-performance compute cluster equipped with NVIDIA H200 GPUs. Specifically, the optimization of the Small variant is distributed across 48 H200 GPUs, while the Large variant requires a cluster of 96 H200 GPUs to handle the increased memory and computational demands. All training operations are performed at a video resolution of $832 \times 480$ to balance visual fidelity with computational throughput during the reinforcement learning rollout phase.

\noindent\textbf{Optimization and Rollout.} We employ the Flow-GRPO-Fast algorithm \citep{liu2025flow} to align the video generation policy. Unlike standard supervised fine-tuning, this online reinforcement learning approach requires generating full video sequences(rollouts) during training to compute the reward signals. For these rollouts, we utilize a denoising schedule to ensure the generated samples are of sufficient quality for the 3D Gaussian Splatting reconstruction process. To stabilize the policy gradient estimation and reduce variance, we configure the GRPO with a group size of $G=8$. The training is scaled with a total batch size distributed across 48 parallel groups, ensuring diverse trajectory sampling for accurate advantage estimation.

\noindent\textbf{Periodic Decoupled Training Strategy.} A critical challenge in reinforcing geometric constraints is the potential suppression of non-rigid dynamics, such as fluid motion or biological deformation. To mitigate this, we implement a Periodic Decoupled Training strategy that alternates between geometric alignment and dynamic preservation. The training process is structured into cycles. In the primary phase, the model is optimized on the full mixed dataset using the complete composite reward function ($R_{3D} + \lambda_{gen}R_{gen}$) to enforce 3D consistency. However, to prevent the model from overfitting to static rigidity, we trigger a \textit{dynamic fine-tuning phase} every 100 training steps. During this interval, we temporarily suspend the 3D-aware reward component $R_{3D}$ and optimize the policy exclusively on the \textit{Dynamic Data Subset}--a curated collection of approximately 500 prompts describing high-entropy scenes (e.g., fire, flowing water, moving crowds). In this phase, the model is guided solely by the general generation reward $R_{gen}$. This periodic relaxation of geometric constraints acts as a regularization mechanism, ensuring that World-R1 retains the generalization power required to synthesize complex, dynamic environments while internalizing the laws of physics.

\section{Dataset Details}
\label{sec:data_details}

To facilitate the training of robust world simulation capabilities, we constructed a proprietary Pure Text Dataset specifically tailored to decouple geometric constraint learning from the biases inherent in existing open-domain video distributions. This dataset, comprising approximately 3,000 unique entries, was synthesized using the advanced instruction-following and creative reasoning capabilities of Gemini~\cite{team2023gemini,team2024gemini}. The data generation process was governed by a hierarchical prompt engineering strategy, ensuring a diverse coverage of visual domains and control complexities. Below, we detail the taxonomy of the camera control primitives and the semantic categories of the dataset, providing the complete list of prompt templates used for training.
\subsection{Generation Pipeline}

The construction of our dataset was automated through a structured generation pipeline leveraging Gemini-3. To ensure the generated text-video pairs effectively disentangle geometric learning from visual bias, we designed a rigorous prompt engineering strategy. This pipeline consists of two core components: a role-playing system instruction for the LLM and a predefined action space for camera control.

\noindent\textbf{Gemini system instruction.}
We instructed the LLM to function as an expert cinematographer, requiring it to synthesize scene descriptions that are not only visually rich but also physically grounded. The system prompt explicitly enforces the pairing of scene geometry with appropriate camera movements (e.g., assigning a \texttt{push\_in} trajectory to a "tunnel" or "corridor" scene to maximize parallax effect). The exact system instruction used in our generation pipeline is presented below:

\begin{center}
\fbox{\begin{minipage}{0.9\textwidth}
\noindent\textbf{Prompt for Dataset Generation}

\noindent\textbf{Role:} You are an expert Cinematographer and 3D Set Designer.
\noindent\textbf{Objective:} Generate a dataset of high-fidelity text-to-video prompts that describe geometrically consistent 3D worlds.
\noindent\textbf{Constraints:}
\begin{itemize}
    \item \textbf{Physical Plausibility:} The scene must follow real-world physics (unless specified as Surrealism). Objects must have defined spatial relationships.
    \item \textbf{Camera-Scene Matching:} The assigned camera movement must logically fit the scene layout (e.g., use 'orbit' for centered objects, 'push\_in' for depth exploration).
    \item \textbf{Diversity:} Cover the following domains: Natural Landscapes, Urban \& Architecture, Micro World, and Fantasy.
\end{itemize}

\noindent\textbf{Action Space (Camera Movements):}
Choose strictly from the following list:
['push\_in', 'pull\_out', 'move\_left', 'move\_right', 'orbit\_left', 'orbit\_right', 'pan\_left', 'pan\_right', 'pull\_left', 'pull\_right', 'fixed']

\noindent\textbf{Output Format (JSON):}
\{
  "prompt": "Full descriptive text...",
  "camera\_logic": "Selected movement",
  "domain": "Category",
  "layout\_type": "Intra/Inter/Static"
\}
\end{minipage}}
\end{center}

\noindent\textbf{Camera trajectory definitions.}
To standardize the action space, we use a comprehensive taxonomy of camera trajectories following WorldScore~\cite{duan2025worldscore}. These are categorized by their topological effect on the scene navigation: \textit{Intra-scene} (exploring within a volume), \textit{Inter-scene} (transitioning across space), and \textit{Static} (stationary observation). The precise definition for each trajectory primitive is detailed below:

\begin{center}
\fbox{\begin{minipage}{0.9\textwidth}
\noindent\textbf{Camera Trajectory Taxonomy}

\noindent\textbf{1. Intra-scene Exploration} \\
\textit{Movements that investigate the depth and 3D structure of a specific subject.}
\begin{itemize}
    \item \textbf{push\_in}: Move forward along the optical axis. (Focus: Depth \& Parallax)
    \item \textbf{orbit\_left}: Revolve counter-clockwise around a focal point. (Focus: 360° Object Consistency)
    \item \textbf{orbit\_right}: Revolve clockwise around a focal point. (Focus: 360° Object Consistency)
\end{itemize}

\noindent\textbf{2. Inter-scene Transition} \\
\textit{Movements that shift the viewport to reveal new environments or expand context.}
\begin{itemize}
    \item \textbf{pull\_out}: Move backward along the optical axis. (Focus: Context Reveal)
    \item \textbf{move\_left}: Lateral truck left. (Focus: Parallax)
    \item \textbf{move\_right}: Lateral truck right. (Focus: Parallax)
    \item \textbf{pan\_left}: Rotate camera yaw left on axis. (Focus: Panoramic View)
    \item \textbf{pan\_right}: Rotate camera yaw right on axis. (Focus: Panoramic View)
\end{itemize}

\noindent\textbf{3. Composite Trajectories} \\
\textit{Complex multi-axis maneuvers testing long-horizon consistency.}
\begin{itemize}
    \item \textbf{pull\_left}: Sequence: \texttt{move\_left} $\rightarrow$ \texttt{pull\_out} $\rightarrow$ \texttt{pan\_left}.
    \item \textbf{pull\_right}: Sequence: \texttt{move\_right} $\rightarrow$ \texttt{pull\_out} $\rightarrow$ \texttt{pan\_right}.
\end{itemize}

\noindent\textbf{4. Static Observation} \\
\textit{Stationary camera to isolate temporal dynamics.}
\begin{itemize}
    \item \textbf{fixed}: No ego-motion. (Focus: Fluid/Particle Dynamics)
\end{itemize}
\end{minipage}}
\end{center}

\subsection{Dataset Taxonomy and Examples}

\subsubsection{Natural Landscapes}

This category focuses on large-scale rigid geometry and natural fluid dynamics. It tests the model's ability to maintain consistency across vast distances and handle organic structures. The prompts are divided into Landforms, Water Features, and Weather \& Time.

\noindent\textbf{Landforms.}
Prompts in this sub-category focus on geological structures and complex terrain traversal.

\begin{center}
\fbox{\begin{minipage}{0.9\textwidth}
\noindent\textbf{Prompts:} \\
A rugged mountain pass covered in scree slopes and hardy alpine flowers. \\
Giant basalt columns forming a hexagonal cliff face by the sea. \\
A lush, hidden valley completely enclosed by towering cliffs. \\
A geothermal area with bubbling mud pools and steaming vents in a rocky landscape. \\
Camera orbit left, transitioning into a pan right. Circling a towering hoodoo, then sweeping across the canyon landscape. \\
Camera pan right, then pull out. A wide view of terraced rice paddies, pulling back to show the entire mountain slope. \\
Camera move left, pull out, then pan right. Starting in a deep gorge, revealing the scale of the cliffs above.
\end{minipage}}
\end{center}

\noindent\textbf{Water features.}
These prompts challenge the model with liquid physics, reflections, and transparency.

\begin{center}
\fbox{\begin{minipage}{0.9\textwidth}
\noindent\textbf{Prompts:} \\
A turquoise lagoon surrounded by white sand beaches and palm trees. \\
Massive ocean waves crashing onto a rocky shore during a storm. \\
Camera move right. Past a frozen waterfall with giant icicles. \\
Camera orbit left. An aerial view circling a series of interlocking lakes. \\
Camera orbit right. Around a large rock formation in the sea. \\
Camera pan left. Scanning the surface of a misty lake. \\
Camera push in, then pan right. Moving towards a misty lake, then looking across the water. \\
Camera pull out, then orbit right. A dramatic reveal of a coastal archway, circling the rock formation. \\
Camera move right, then push in. Along a bubbling brook, then focusing on a small fish.
\end{minipage}}
\end{center}

\noindent\textbf{Weather \& time.}
These entries evaluate atmospheric rendering and dynamic lighting conditions.

\begin{center}
\fbox{\begin{minipage}{0.9\textwidth}
\noindent\textbf{Prompts:} \\
A dramatic cloudscape at sunset, with towering cumulus clouds catching the light. \\
A peaceful, starry night over a calm mountain lake, with stars reflected in the water. \\
A foggy coastline with the sound of a foghorn in the distance. \\
Camera fixed. A dramatic view of a sunset reflection on wet sand. \\
Camera push in. Into the pure blackness of a moonless night in the wilderness. \\
Camera pull out. Revealing a hazy jungle canopy at sunrise. \\
Camera move left. Watch lightning bolts strike during a night storm. \\
Camera move right. Across a desaturated landscape on an overcast day. \\
Camera orbit left. Around a sunrise with sun pillars. \\
Camera orbit right. Circling trees with leaves blowing on a windy autumn day. \\
Camera pan left. Across a twilight desert sky with color gradients. \\
Camera pan right. Following falling snow in a forest at night. \\
Camera fixed. A view of crepuscular rays over the ocean. \\
Camera push in. Towards a crisp, clean landscape after a rainstorm. \\
Camera pull out. Revealing low-lying fog blanketing a field at dawn. \\
Camera pan left, then push in. Scanning a hazy jungle sunrise before focusing on the sun. \\
Camera move left, pull out, then pan left. Starting with lightning striking, revealing the thunderstorm at night. \\
Camera push in, then pan right. Moving towards an overcast landscape, then scanning the flat light. \\
Camera pull out, then orbit right. A dramatic reveal of a sunrise with sun pillars, circling the light phenomenon. \\
Camera move right, then push in. Along windy autumn trees, focusing on a blowing leaf.
\end{minipage}}
\end{center}

\subsubsection{Urban and Architectural}

The Urban and Architectural domain emphasizes strict perspective correctness, vanishing points, and the preservation of straight lines.

\noindent\textbf{Urban landscapes.}
\begin{center}
\fbox{\begin{minipage}{0.9\textwidth}
\noindent\textbf{Prompts:} \\
A futuristic cyberpunk city street drenched in neon rain at night. \\
A charming cobblestone street in a medieval European town at sunrise. \\
The skyline of a modern metropolis reflecting in a calm bay at dusk. \\
The historic Bund in Shanghai illuminated with golden lights at night. \\
A post-apocalyptic city ruin reclaimed by nature, with vines growing on skyscrapers. \\
A busy intersection in Tokyo filled with pedestrians and colorful signs. \\
Camera pull out, then orbit left. A dramatic reveal of a floating city, circling the platform. \\
Camera move right, then push in. Along a rainy street, then focusing on a reflection. \\
Camera orbit left, transitioning into a pan right. Circling a town square, then sweeping the crowd. \\
Camera pan right, then pull out. A wide view of traffic, pulling back to show the gridlock. \\
Camera move left, pull out, then pan right. Starting in a garden, revealing the suburban neighborhood.
\end{minipage}}
\end{center}

\noindent\textbf{Indoor spaces.}
\begin{center}
\fbox{\begin{minipage}{0.9\textwidth}
\noindent\textbf{Prompts:} \\
Camera pull out. Revealing a vast warehouse interior. \\
Camera move left. Along the lockers in a school hallway. \\
Camera move right. Past gaming machines in a vintage arcade. \\
Camera orbit left. Circling a plant in a greenhouse. \\
Camera orbit right. Around a table in a conference room. \\
Camera pan left. Across a luxury hotel lobby. \\
Camera pan right. Over the water in an indoor pool. \\
Camera fixed. A view of a rustic kitchen. \\
Camera push in. Into a dark corridor in an abandoned hospital. \\
Camera pull out. Revealing a packed casino floor. \\
Camera move left. Along the racks in a walk-in closet.
\end{minipage}}
\end{center}

\noindent\textbf{Infrastructure.}
\begin{center}
\fbox{\begin{minipage}{0.9\textwidth}
\noindent\textbf{Prompts:} \\
An empty, futuristic metro station with polished floors and blue lighting. \\
A lonely gas station on a desert highway illuminated by fluorescent lights. \\
A massive suspension bridge spanning a foggy strait. \\
Camera move left, then push in. Along a train platform, then approaching the track. \\
Camera orbit right, transitioning into a pull out. Circling a gas pump, then pulling back to show the station. \\
Camera pan left, then push in. Scanning a factory floor, then focusing on a machine. \\
Camera move right, pull out, then pan right. Starting at a suitcase, revealing the airport terminal.
\end{minipage}}
\end{center}

\subsubsection{Micro and Still Life}

Designed to evaluate depth-of-field handling and texture fidelity, this category focuses on small-scale objects and macro observation.

\noindent\textbf{Desktop still life.}
\begin{center}
\fbox{\begin{minipage}{0.9\textwidth}
\noindent\textbf{Prompts:} \\
Camera pull out. Revealing a messy artist's desk. \\
Camera move left. Along the keys of a mechanical keyboard. \\
Camera move right. Past a row of antique books. \\
Camera orbit left. Circling a vase of flowers. \\
Camera orbit right. Around a burning candle flame. \\
Camera pan left. Scanning a table of food. \\
Camera pan right. Across a blueprint on a drafting table. \\
Camera fixed. A view of an hourglass with falling sand. \\
Camera push in. Towards the nib of a fountain pen. \\
Camera pull out. From a macro of a coin to a pile of money. \\
Camera move left. Along the spine of a leather notebook. \\
Camera move right. Past spools of colorful thread. \\
Camera push in, then orbit left. Approaching a feather quill, then circling the inkwell. \\
Camera pull out, then pan left. Revealing a record player from the needle. \\
Camera move right, then push in. Along a coaster, then focusing on a water ring. \\
Camera orbit right, then pull out. Circling a chocolate, then showing the box. \\
Camera pan left, then push in. Scanning a newspaper, then focusing on a headline. \\
Camera move left, pull out, then pan left. Starting at a screw, revealing the glasses. \\
Camera push in, then pan right. Moving towards a matcha whisk, then looking at the foam.

\end{minipage}}
\end{center}

\noindent\textbf{Micro world.}
\begin{center}
\fbox{\begin{minipage}{0.9\textwidth}
\noindent\textbf{Prompts:} \\
A close-up of a butterfly wing, revealing the mosaic of tiny scales. \\
The fuzzy texture of mold growing on a piece of bread. \\
A microscopic view of blood cells flowing through a capillary. \\
The jagged landscape of a vinyl record groove viewed under a microscope. \\
A grain of sand that looks like a jewel stone when magnified. \\
The delicate stamens of a hibiscus flower covered in yellow pollen. \\
A close-up of bubbles rising in a glass of carbonated soda. \\
The rough, cratered surface of a match stick head before ignition. \\
A spider web covered in tiny water droplets, glistening in the sun. \\
The fibers of a piece of paper, looking like a tangled forest. \\
A macro view of human iris, showing the intricate muscle patterns. \\
The sharp, serrated edge of a kitchen knife blade. \\
A close-up of moss sporophytes looking like an alien forest. \\
The crystalline structure of sugar granules spilled on a dark surface. \\
A micro view of a feather, showing the barbs and barbules. \\
The glowing filament of a lightbulb, vibrating with heat. \\
Camera fixed. A view of moss sporophytes. \\
Camera push in. Towards the nucleus of a cell. \\
Camera pull out. From a facet to a gemstone view. \\
Camera move left. Along a crack in glass. \\
Camera move right. Past glowing embers. \\
Camera orbit left. Circling a pollen grain. \\
Camera orbit right. Around a mosquito. \\
Camera pan left. Across a bird egg shell. \\
Camera pan right. Over a CD surface.

\end{minipage}}
\end{center}

\noindent\textbf{Material representation.}
\begin{center}
\fbox{\begin{minipage}{0.9\textwidth}
\noindent\textbf{Prompts:} \\
The iridescent surface of a soap bubble swirling with colors. \\
A patch of soft, white animal fur, detailed and fluffy. \\
Rusty iron metal surface with flaking orange and brown corrosion. \\
Translucent jade stone with internal cloudy patterns. \\
A wet cobblestone street reflecting streetlights at night. \\
A piece of honeycomb dripping with thick, golden honey. \\
The rough, porous texture of a lava rock. \\
Silky satin fabric flowing like liquid in a gentle breeze. \\
A close-up of carbon fiber weave, black and geometric. \\
The sparkling surface of fresh snow in sunlight. \\
A pool of crude oil, thick, black, and viscous. \\
Old, cracked leather texture on an antique armchair. \\
A mosaic of stained glass pieces, glowing with transmitted light. \\
The intricate pattern of a snake's scales. \\
A close-up of a woven wicker basket. \\
Molten lava flowing and cooling into black rock. \\
Camera pull out. Revealing a tire tread. \\
Camera move left. Along a copper pipe. \\
Camera move right. Past a jelly cube. \\
Camera orbit left. Circling a tennis ball. \\
Camera move right, then push in. Along a leather crack, then focusing on the texture. \\
Camera orbit left, transitioning into a pan right. Circling a stained glass piece, then sweeping the window. \\
Camera pan right, then pull out. A wide view of wet cobblestones, pulling back to the street. \\
Camera move left, pull out, then pan right. Starting at a sequin, revealing the dress.
\end{minipage}}
\end{center}

\subsubsection{Fantasy and Surrealism}

This category challenges the model's generalization capabilities by introducing non-Euclidean geometries and physics-defying structures.

\begin{center}
\fbox{\begin{minipage}{0.9\textwidth}
\noindent\textbf{Prompts:} \\
A floating library orbiting a small star. \\
A city where gravity shifts between districts. \\
A dreamlike ruin suspended in a crystal sphere. \\
A surreal canyon filled with echoing light. \\
A floating island with an endless spiral tower. \\
A dream world where the ground reflects the sky. \\
A surreal city constructed from ancient stone faces. \\
A floating garden drifting through a pastel sky. \\
A dreamlike shoreline where water turns into glass. \\
A surreal city half submerged in glowing mist. \\
A floating citadel guarded by silent stone figures. \\
A dreamscape where mountains fold like paper. \\
A surreal field filled with hovering monoliths. \\
A floating city beneath an aurora-filled sky. \\
A dream world where light casts solid shadows. \\
A surreal bridge stretching across a starless void. \\
A floating village supported by giant birds. \\
A dreamlike forest where the air shimmers. \\
A surreal city built inside a colossal hourglass. \\
A floating island with inverted waterfalls. \\
A dreamscape where the moon rests on the horizon. \\
A surreal canyon illuminated by glowing symbols. \\
Camera move right. A dreamscape where shadows form buildings. \\
Camera orbit left. A surreal forest glowing softly at night. \\
Camera orbit right. A floating tower surrounded by glowing fog. \\
Camera pan left. A dream world where gravity reverses at dawn. \\
Camera pan right. A surreal bridge stretching across a void. \\
Camera push in. A floating ruin wrapped in glowing vines. \\
Camera pull out. A dreamscape where the sky shatters softly. \\
Camera move left. A surreal city illuminated by floating crystals. \\
Camera move right. A floating island drifting through mist. \\
Camera orbit left. A dream world where space folds inward. \\
Camera push in, then orbit left. A floating island with inverted waterfalls. \\
Camera pull out, then pan right. A surreal city built on clock gears. \\
Camera move left, then push in. A dreamscape where shadows float freely. \\
Camera move right, then orbit right. A floating palace drifting through fog. \\
Camera push in, then pan left. A surreal canyon filled with glowing glyphs. \\
Camera pull out, then orbit left. A floating observatory surrounded by stars. \\
Camera move left, then pan left. A dream world where geometry bends. \\
Camera move right, then pan right. A surreal forest of bioluminescent trees. \\
Camera push in, then orbit right. A floating city glowing at eternal dusk. \\
Camera pull out, then pan left. A dreamscape where the horizon curves. \\
Camera move left, then push in. A surreal island chained to the moon. \\
Camera move right, then push in. A floating ruin wrapped in glowing vines. \\
Camera push in, then pull out. A dream world where buildings breathe softly. \\
Camera orbit left, then pan left. A surreal valley bathed in violet light. \\
Camera orbit right, then pan right. A floating city drifting apart.
\end{minipage}}
\end{center}

\subsubsection{Artistic Styles}

This category ensures that the 3D constraints do not degrade the aesthetic diversity of the foundation model, requiring distinct stylized renderings rather than photorealism.

\begin{center}
\fbox{\begin{minipage}{0.9\textwidth}
\noindent\textbf{Prompts:} \\
A campfire in snowy woods, watercolor painting \\
A street musician performing at night, black and white photography \\
A flickering neon sign in fog, cyberpunk illustration \\
A castle made of glass, fantasy illustration \\
Camera move right. A sunflower field under swirling sky, Van Gogh style \\
Camera orbit left. A cracked marble statue repaired with gold, kintsugi art \\
Camera orbit right. A glowing crystal floating in darkness, fantasy digital art \\
Camera push in. A girl holding a lantern, watercolor painting \\
Camera pull out. An infinite library interior, surrealism \\
Camera pan left. A snowy village at dusk, storybook illustration \\
Camera pan right. A futuristic skyline at night, matte painting \\
Camera move left. A cat sleeping on stacked books, pastel illustration \\
Camera move right. A mechanical bird on a branch, steampunk illustration \\
Camera orbit left. A floating violin, surreal illustration
\end{minipage}}
\end{center}

\subsection{Dynamic Data Subset}

This subset focuses on high-entropy scenes and deformable objects to mitigate the suppression of non-rigid dynamics during 3D alignment.

\begin{center}
\fbox{\begin{minipage}{0.9\textwidth}
\noindent\textbf{Prompts:} \\
A massive waterfall cascading down a cliff with mist rising. \\
Camera push in. A lion roaring with its mane shaking in the wind. \\
Camera orbit left, then pull out. A futuristic car transforming into a robot. \\
Camera pan right. A cyclist sprinting towards the finish line. \\
Camera fixed. Raindrops splashing into a puddle, creating ripples. \\
Camera move left, pull out, then pan left. A spaceship launching with thrusters igniting. \\
A chef tossing vegetables in a flaming wok. \\
Camera orbit right. A tornado tearing through a wooden barn. \\
Camera pull out, then pan right. Hundreds of lanterns floating up into the night sky. \\
Camera move right. A subway train rushing past the platform. \\
Camera push in, then pan left. A flower blooming rapidly in time-lapse. \\
A glass shattering on the floor in slow motion. \\
Camera pan left. Soldiers marching in synchronization across a dusty field. \\
Camera orbit left. A breakdancer spinning on their head. \\
Camera move right, pull out, then pan right. A whale breaching the ocean surface. \\
A volcano erupting with lava flowing down the side.
\end{minipage}}
\end{center}

\section{User Study Details}
\label{sec:user_study}

To comprehensively evaluate the perceptual quality of World-R1 compared to baseline foundation models, we conducted a blind subjective user study. While quantitative metrics provide objective measures of reconstruction, human evaluation remains the gold standard for assessing generative video quality, particularly for semantic adherence and visual plausibility.

\subsection{Experimental Setup}

\noindent\textbf{Methodology.} We adopted a Two-Alternative Forced Choice (2AFC) protocol. For each evaluation instance, participants were presented with two video clips played side-by-side: one generated by our \method model and the other by the corresponding base model (Wan 2.1). The position of the videos (left vs. right) was randomized for every query to eliminate positional bias. The study was strictly double-blind; neither the participants nor the administrators knew which model produced which video during the evaluation phase.

\noindent\textbf{Model Comparisons.} To ensure a fair comparison, we matched models by parameter size:
\begin{itemize}
    \item \textbf{Small Variant:} \textit{World-R1-Small} was compared against \textit{Wan2.1-T2V-1.3B}.
    \item \textbf{Large Variant:} \textit{World-R1-Large} was compared against \textit{Wan2.1-T2V-14B}.
\end{itemize}

\noindent\textbf{Participants.} The study involved a total of 25 participants. The pool consisted of evaluators with varying degrees of familiarity with generative AI, ensuring the results reflect both expert critique and general user preference.

\subsection{Evaluation Dataset}

We curated a fixed evaluation set of \textbf{30 complex prompts} designed to stress-test the models' capabilities in world modeling. The dataset was distinct from the training set and covered a diverse range of scenarios.

\subsection{Evaluation Criteria}

Participants were instructed to watch the video pairs and select the superior model based on three independent criteria. We provided specific definitions for each criterion to ensure consistent grading:

\begin{enumerate}
    \item \textbf{Geometric Consistency:}
    \textit{"Which video better maintains the structural integrity of the physical world?"}
    \begin{itemize}
        \item \textbf{Look for:} Objects that remain solid (no morphing/vanishing), background elements that stay fixed relative to the camera movement, and correct perspective changes (parallax).
        \item \textbf{Penalize:} Walls that warp, objects that float or disappear when the camera moves, and "dream-like" transitions in rigid environments.
    \end{itemize}

    \item \textbf{Camera Control Accuracy:}
    \textit{"Which video more accurately executes the camera movement described in the text prompt?"}
    \begin{itemize}
        \item \textbf{Look for:} Strict adherence to the directional command (e.g., if the prompt says "orbit," the camera must circle the object, not just pan or zoom).
        \item \textbf{Penalize:} Errant camera drift, static cameras when motion is requested, or wrong movement directions.
    \end{itemize}

    \item \textbf{Overall Visual Quality:}
    \textit{"Which video do you prefer overall?"}
    \begin{itemize}
        \item \textbf{Look for:} High image fidelity, aesthetic appeal, lack of visual artifacts (blurriness, noise), and general coherence.
        \item \textbf{Note:} This metric captures the trade-off between strict adherence to physics and artistic generation quality.
    \end{itemize}
\end{enumerate}

\subsection{Detailed Results}

We aggregated the responses across all participants and prompts. The Win Rate represents the percentage of instances where \method was selected as superior to the baseline \textit{Wan 2.1}.

\begin{table}[h]
\centering
\caption{User Study Results (Win Rate of World-R1 vs. Wan 2.1). The results demonstrate a decisive preference for World-R1 in tasks requiring physical reasoning and control.}
\label{tab:user_study}
\begin{tabular}{@{}p{0.29\textwidth}p{0.15\textwidth}p{0.46\textwidth}@{}}
\toprule
\noindent\textbf{Metric} & \textbf{Win Rate (\%)} & \textbf{Analysis} \\
\midrule
Geometric Consistency & \textbf{92\%} & Superior structural stability; significant reduction in hallucination. \\
Camera Control Accuracy & \textbf{76\%} & Better instruction following for complex trajectories. \\
Overall Preference & \textbf{86\%} & Users prefer the aligned, consistent videos despite the constraints. \\
\bottomrule
\end{tabular}
\end{table}

The results indicate that while the baseline models are capable of generating visually pleasing individual frames, they frequently fail to maintain 3D consistency over time. \method's dominant performance in Geometric Consistency ($92\%$) validates the effectiveness of our RL-based alignment strategy in teaching the model latent 3D constraints.

\subsection{Metric Validation Study}

To further verify that our automatic 3D-consistency metric reflects human perception rather than artifacts of the reconstruction pipeline, we conducted an additional metric-validation user study. This study involved 20 participants and 30 randomized video pairs. For each pair, participants selected the video with stronger perceived 3D consistency, and we compared the majority human preference with the ranking induced by our automatic metric. As shown in \cref{tab:metric_user_study}, human preference aligns with the metric ranking in 91.17\% of comparisons, confirming that the metric is a reliable perceptual proxy for geometric consistency.

\begin{table}[h]
\centering
\caption{Metric-validation user study. Agreement measures how often majority human preference on 3D consistency matches the automatic metric ranking.}
\label{tab:metric_user_study}
\begin{tabular}{lccc}
\toprule
\noindent\textbf{Study} & \textbf{Participants} & \textbf{Video Pairs} & \textbf{Agreement} $\uparrow$ \\
\midrule
3D-consistency metric validation & 20 & 30 & \textbf{91.17\%} \\
\bottomrule
\end{tabular}
\end{table}

\section{Additional Experimental Analyses}
\label{sec:additional_exp}

This section provides additional experiments motivated by reviewer feedback. We place these analyses in the appendix to keep the main paper focused while making the empirical validation more comprehensive.

\subsection{Camera-Control Accuracy}

We further evaluate camera-control accuracy using RotErr, TransErr, and CamMC under a reference-video trajectory protocol. Despite being trained as a post-training alignment method rather than a dedicated camera-control architecture, \method achieves camera-control accuracy competitive with specialized methods while preserving stronger general video quality.

\begin{table}[h]
\centering
\caption{Camera-control evaluation. Lower RotErr, TransErr, and CamMC are better.}
\label{tab:camera_control}
\begin{tabular}{lccc}
\toprule
\noindent\textbf{Method} & \textbf{RotErr} $\downarrow$ & \textbf{TransErr} $\downarrow$ & \textbf{CamMC} $\downarrow$ \\
\midrule
ReCamMaster~\cite{bai2025recammaster} & 1.53 & 3.12 & 4.17 \\
TrajectoryCrafter~\cite{yu2025trajectorycrafter} & 3.08 & 7.46 & 10.22 \\
CamCloneMaster~\cite{luo2025camclonemaster} & 1.36 & 2.02 & 3.05 \\
Wan2.1-T2V-1.3B~\cite{wan2025wan} & 9.29 & 62.94 & 66.21 \\
Wan2.1-T2V-14B~\cite{wan2025wan} & 17.01 & 60.90 & 70.55 \\
\noindent\textbf{\method-Small} & 1.50 & 2.76 & 3.39 \\
\noindent\textbf{\method-Large} & \textbf{1.21} & \textbf{1.30} & \textbf{2.95} \\
\bottomrule
\end{tabular}
\end{table}

\subsection{Reconstruction-Independent Multi-View Consistency}

Our main 3D-consistency metric uses a shared 3DGS reconstruction pipeline for all compared methods, which ensures a fair comparison but may still inherit biases from the reconstruction model. To verify that the improvements do not merely come from the reconstruction process, we additionally evaluate Multi-View Consistency Score (MVCS) following GeoVideo~\cite{bai2025geovideo}, a reconstruction-independent metric that directly measures cross-view agreement from generated videos. As shown in \cref{tab:mvcs}, \method improves MVCS for both the small and large backbones, confirming that the gains stem from the generated videos rather than artifacts of 3DGS optimization.

\begin{table}[h]
\centering
\caption{Reconstruction-independent multi-view consistency evaluation. Higher MVCS indicates stronger cross-view agreement.}
\label{tab:mvcs}
\begin{tabular}{lc}
\toprule
\noindent\textbf{Method} & \textbf{MVCS} $\uparrow$ \\
\midrule
Wan2.1-T2V-1.3B~\cite{wan2025wan} & 0.974 \\
\noindent\textbf{\method-Small (Ours)} & \textbf{0.989} \\
Wan2.1-T2V-14B~\cite{wan2025wan} & 0.963 \\
\noindent\textbf{\method-Large (Ours)} & \textbf{0.993} \\
\bottomrule
\end{tabular}
\end{table}

\subsection{Dataset Scaling}

We evaluate the effect of increasing the size of the pure text post-training dataset. \cref{tab:data_scaling} shows a consistent positive trend from 1K to 3K prompts across both 3D consistency and general video quality. The result indicates that \method is data-efficient, already producing strong gains with 3K prompts, and suggests that further scaling of prompt generation could yield additional improvements.

\begin{table}[h]
\centering
\caption{Dataset scaling study on \method-Small. Increasing prompt data improves both geometric consistency and general video quality.}
\label{tab:data_scaling}
\begin{tabular}{lcccc}
\toprule
\noindent\textbf{Data Size} & \textbf{PSNR} $\uparrow$ & \textbf{SSIM} $\uparrow$ & \textbf{LPIPS} $\downarrow$ & \textbf{VBench AVG} $\uparrow$ \\
\midrule
1K & 25.82 & 0.812 & 0.258 & 83.23 \\
2K & 26.54 & 0.839 & 0.223 & 84.76 \\
3K & \textbf{27.63} & \textbf{0.858} & \textbf{0.201} & \textbf{85.21} \\
\bottomrule
\end{tabular}
\end{table}

\subsection{Long-Video Generalization}

Although training is conducted on short clips, \method generalizes to longer generations. We evaluate 121-frame videos in \cref{tab:long_video}. \method-Large substantially improves all 3D-consistency metrics over the Wan2.1-T2V-14B backbone, suggesting that the learned geometric alignment transfers beyond the training horizon.

\begin{table}[h]
\centering
\caption{Long-video evaluation on 121-frame generations.}
\label{tab:long_video}
\begin{tabular}{lccc}
\toprule
\noindent\textbf{Method} & \textbf{PSNR} $\uparrow$ & \textbf{SSIM} $\uparrow$ & \textbf{LPIPS} $\downarrow$ \\
\midrule
Wan2.1-T2V-14B~\cite{wan2025wan} & 18.32 & 0.558 & 0.534 \\
\noindent\textbf{\method-Large (Ours)} & \textbf{26.32} & \textbf{0.828} & \textbf{0.257} \\
\bottomrule
\end{tabular}
\end{table}

\subsection{Scene-Complexity Breakdown}

To evaluate challenging settings more systematically, we categorize the test prompts by scene type and report per-category results in \cref{tab:scene_breakdown}. \method-Small consistently improves over the Wan2.1-T2V-1.3B backbone across static scenes, single-object dynamics, multi-object dynamics, non-rigid motion, and selected long-horizon dynamics. The largest absolute challenge remains long-horizon and non-rigid generation, where final quality is partly limited by the base model's compositional and motion-generation capacity; nevertheless, the improvements show that RL-based 3D alignment remains effective in these regimes.

\begin{table}[h]
\centering
\caption{Per-category results by scene complexity. ``N'' denotes the percentage of prompts in each category. Long-horizon dynamics are selected from the preceding categories.}
\label{tab:scene_breakdown}
\resizebox{\textwidth}{!}{%
\begin{tabular}{lllcccc}
\toprule
\noindent\textbf{Scene Type} & \textbf{N} & \textbf{Method} & \textbf{PSNR} $\uparrow$ & \textbf{SSIM} $\uparrow$ & \textbf{LPIPS} $\downarrow$ & \textbf{MVCS} $\uparrow$ \\
\midrule
Static Scene & 30.11\% & Wan2.1-1.3B & 20.14 & 0.632 & 0.389 & 0.981 \\
Static Scene & 30.11\% & \textbf{\method-Small} & \textbf{30.52} & \textbf{0.912} & \textbf{0.142} & \textbf{0.994} \\
Single-obj Dynamic & 29.03\% & Wan2.1-1.3B & 17.86 & 0.563 & 0.452 & 0.976 \\
Single-obj Dynamic & 29.03\% & \textbf{\method-Small} & \textbf{28.17} & \textbf{0.869} & \textbf{0.189} & \textbf{0.991} \\
Multi-obj Dynamic & 21.51\% & Wan2.1-1.3B & 15.23 & 0.487 & 0.528 & 0.968 \\
Multi-obj Dynamic & 21.51\% & \textbf{\method-Small} & \textbf{25.41} & \textbf{0.812} & \textbf{0.248} & \textbf{0.985} \\
Non-rigid Motion & 19.35\% & Wan2.1-1.3B & 14.58 & 0.462 & 0.548 & 0.965 \\
Non-rigid Motion & 19.35\% & \textbf{\method-Small} & \textbf{24.73} & \textbf{0.793} & \textbf{0.267} & \textbf{0.982} \\
Long-horizon Dynamics & 12.89\% & Wan2.1-1.3B & 12.53 & 0.382 & 0.683 & 0.951 \\
Long-horizon Dynamics & 12.89\% & \textbf{\method-Small} & \textbf{23.59} & \textbf{0.781} & \textbf{0.299} & \textbf{0.974} \\
\bottomrule
\end{tabular}}
\end{table}

\subsection{Comparison with 3D-Aware Generation Methods}

We further consolidate comparisons with publicly available 3D-aware video generation methods. These methods fall into two categories: camera-control methods that introduce explicit control modules, and 3D-conditioned methods that rely on architectural modifications or image-to-video pipelines. In contrast, \method is a text-to-video post-training framework requiring no inference-time architectural changes. As shown in \cref{tab:3d_aware_comparison}, \method achieves the strongest 3D-consistency metrics while preserving competitive general video quality. These results indicate that our post-training paradigm is complementary to architecture-level 3D-aware designs and can potentially be applied on top of them.

\begin{table}[h]
\centering
\caption{Consolidated comparison with 3D-aware and camera-control generation methods. The 3D-conditioned baselines are evaluated in their image-to-video setting, whereas \method is evaluated as a text-to-video model on general scenes.}
\label{tab:3d_aware_comparison}
\resizebox{\textwidth}{!}{%
\begin{tabular}{llcccccccc}
\toprule
\noindent\textbf{Type} & \textbf{Method} & \textbf{PSNR} $\uparrow$ & \textbf{SSIM} $\uparrow$ & \textbf{LPIPS} $\downarrow$ & \textbf{MVCS} $\uparrow$ & \textbf{Aesthetic} $\uparrow$ & \textbf{BG Cons.} $\uparrow$ & \textbf{Subject Cons.} $\uparrow$ & \textbf{Motion Smooth.} $\uparrow$ \\
\midrule
3D-Cond. & ViewCrafter~\cite{yu2024viewcrafter} & 23.15 & 0.724 & 0.291 & 0.979 & 55.52 & 92.09 & 94.25 & 97.86 \\
3D-Cond. & Voyager~\cite{huang2025voyager} & 21.38 & 0.678 & 0.334 & 0.975 & 49.80 & 92.31 & 91.55 & 99.39 \\
3D-Cond. & FlashWorld~\cite{li2025flashworld} & 22.46 & 0.702 & 0.312 & 0.977 & 53.72 & 91.88 & 94.44 & 98.81 \\
3D-Cond. & VerseCrafter~\cite{zheng2026versecrafter} & 23.82 & 0.748 & 0.268 & 0.981 & 54.78 & 94.88 & 95.55 & 97.62 \\
Cam. Ctrl. & GCD~\cite{van2024generative} & 18.26 & 0.582 & 0.438 & 0.966 & 38.21 & 92.00 & 88.94 & 98.37 \\
Cam. Ctrl. & Traj.-Attn.~\cite{xiao2024trajectory} & 18.87 & 0.598 & 0.421 & 0.969 & 38.50 & 92.83 & 90.60 & 98.21 \\
Cam. Ctrl. & DAS~\cite{gu2025diffusion} & 19.42 & 0.618 & 0.398 & 0.971 & 39.86 & 92.03 & 90.34 & 99.14 \\
Cam. Ctrl. & ReCamMaster~\cite{bai2025recammaster} & 20.58 & 0.653 & 0.368 & 0.975 & 42.70 & 93.83 & 92.05 & 99.28 \\
Foundation & Wan2.1-1.3B~\cite{wan2025wan} & 17.40 & 0.550 & 0.467 & 0.974 & 62.43 & 97.29 & 96.34 & 97.44 \\
Ours & \textbf{\method-Small} & \textbf{27.63} & \textbf{0.858} & \textbf{0.201} & \textbf{0.989} & \textbf{65.74} & 96.67 & \textbf{97.58} & 98.55 \\
\bottomrule
\end{tabular}}
\end{table}

\subsection{Reward Hacking Analysis}
\label{sec:reward_hacking}

A potential failure mode of RL-based alignment is reward hacking: if the model optimizes only a narrow 3D metric, it may learn degenerate shortcuts such as producing near-static clips that are easy to reconstruct but do not follow the desired camera motion or preserve dynamic content. Our design mitigates this issue in three ways. First, the composite reward combines meta-view plausibility, reconstruction fidelity, trajectory alignment, and general video-generation quality, so no single metric can dominate the optimization. Second, the trajectory term penalizes static outputs when a non-trivial camera trajectory is requested. Third, periodic decoupled training on dynamic prompts prevents the model from collapsing into overly rigid generation.

\cref{tab:reward_ablation} reports component-wise reward ablations. Removing any reward component degrades performance, and the full pipeline achieves the best overall trade-off across 3D consistency and VBench quality. In particular, removing $\mathcal{S}_{traj}$ weakens trajectory adherence, while removing $\mathcal{S}_{meta}$ or $\mathcal{S}_{recon}$ reduces geometric consistency. \cref{tab:training_ablation} further analyzes the core training and conditioning components together with the decomposed reward signals. Removing periodic decoupled training can improve reconstruction-style scores by pushing the model toward overly rigid or near-static solutions, but it degrades VBench quality. Removing the 3D-aware reward preserves general video quality but fails to regularize geometry. These results support that the reward components act as complementary safeguards against reward exploitation rather than redundant terms.

\begin{table}[h]
\centering
\caption{Reward component ablation on \method-Small. Each component contributes to the final balance between geometric consistency and general video quality.}
\label{tab:reward_ablation}
\begin{tabular}{lcccc}
\toprule
\noindent\textbf{Variant} & \textbf{PSNR} $\uparrow$ & \textbf{SSIM} $\uparrow$ & \textbf{LPIPS} $\downarrow$ & \textbf{VBench AVG} $\uparrow$ \\
\midrule
Full pipeline (\method-Small) & \textbf{27.63} & \textbf{0.858} & \textbf{0.201} & \textbf{85.21} \\
w/o $\mathcal{S}_{meta}$ & 26.91 & 0.841 & 0.218 & 83.67 \\
w/o $\mathcal{S}_{recon}$ & 25.14 & 0.798 & 0.271 & 84.35 \\
w/o $\mathcal{S}_{traj}$ & 26.27 & 0.829 & 0.237 & 84.53 \\
\bottomrule
\end{tabular}
\end{table}

\begin{table}[h]
\centering
\caption{Training and conditioning ablation with decomposed reward scores. Dashes indicate disabled reward terms.}
\label{tab:training_ablation}
\resizebox{\textwidth}{!}{%
\begin{tabular}{lcccccccc}
\toprule
\noindent\textbf{Variant} & \textbf{PSNR} $\uparrow$ & \textbf{SSIM} $\uparrow$ & \textbf{LPIPS} $\downarrow$ & \textbf{VBench AVG} $\uparrow$ & $\mathcal{S}_{recon}$ $\uparrow$ & $\mathcal{S}_{traj}$ $\uparrow$ & $\mathcal{S}_{meta}$ $\uparrow$ & $\mathcal{S}_{gen}$ $\uparrow$ \\
\midrule
Full & \textbf{27.63} & 0.858 & 0.201 & \textbf{85.21} & 0.342 & 0.296 & \textbf{0.307} & \textbf{0.37} \\
w/o noise wrapping & 24.46 & 0.745 & 0.298 & 76.39 & 0.312 & 0.213 & 0.275 & -0.42 \\
w/o periodic decoupled training & 27.89 & \textbf{0.898} & \textbf{0.192} & 82.64 & 0.348 & \textbf{0.310} & 0.298 & 0.18 \\
w/o 3D-aware reward & 18.93 & 0.502 & 0.496 & 84.96 & -- & -- & -- & 0.33 \\
w/o general reward & 27.57 & 0.849 & 0.206 & 83.44 & \textbf{0.388} & 0.231 & 0.305 & -- \\
\bottomrule
\end{tabular}}
\end{table}

\section{More Visualizations}
\label{sec:more_vis}

For video results, please refer to the video in supplementary material.

\begin{figure}[t]
  \centering
  \includegraphics[width=\textwidth]{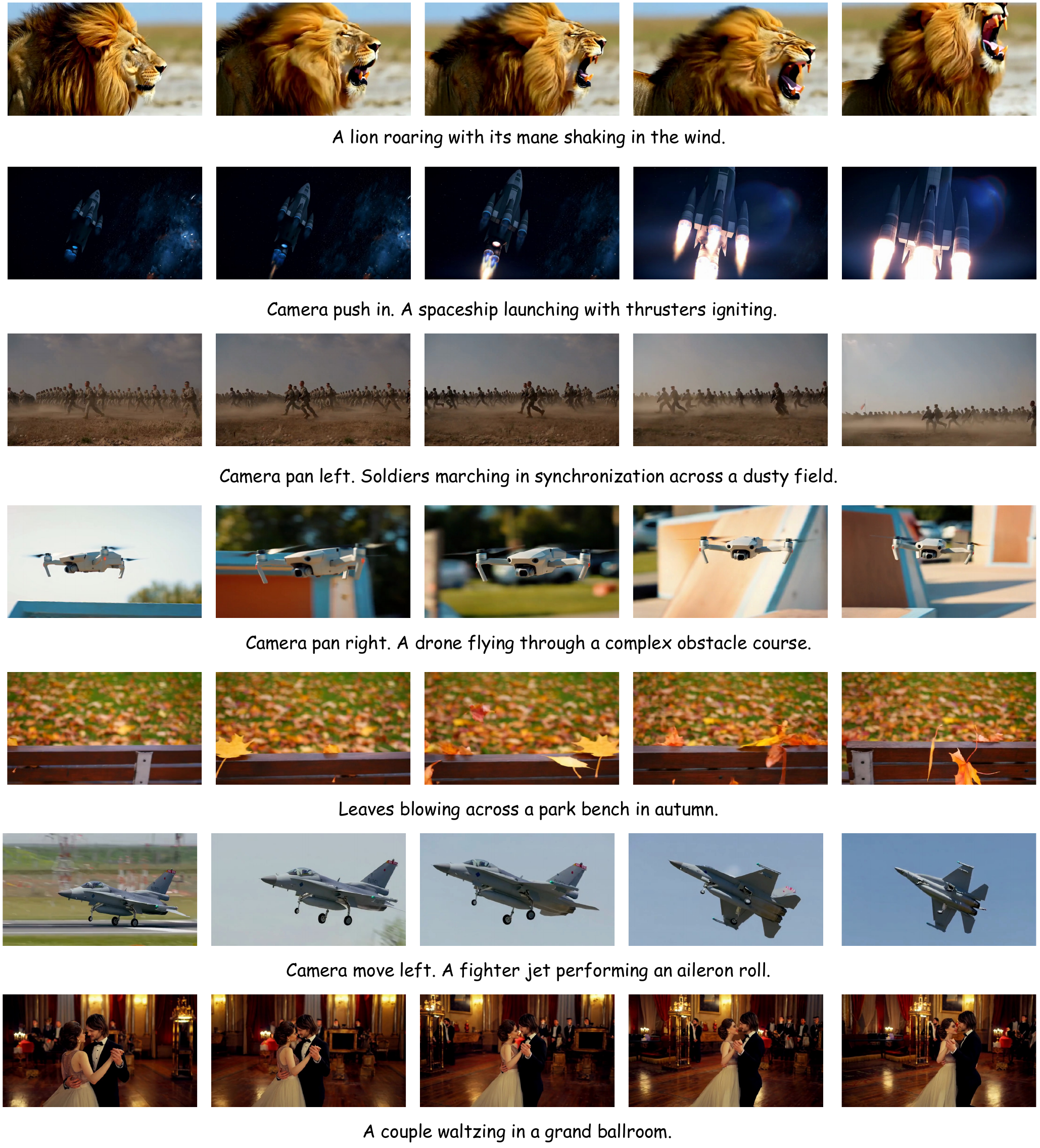}
  \caption{\textbf{Qualitative results on dynamic scenes.} We visualize samples from World-R1 on prompts requiring high temporal dynamics. Despite being trained with strict 3D consistency rewards, the model preserves the natural motion of non-rigid elements, demonstrating the effectiveness of our periodic decoupled training strategy.}
  \label{fig:dynamic_qualitative} 
\end{figure}

\begin{figure}[t]
  \centering
  \includegraphics[width=\textwidth]{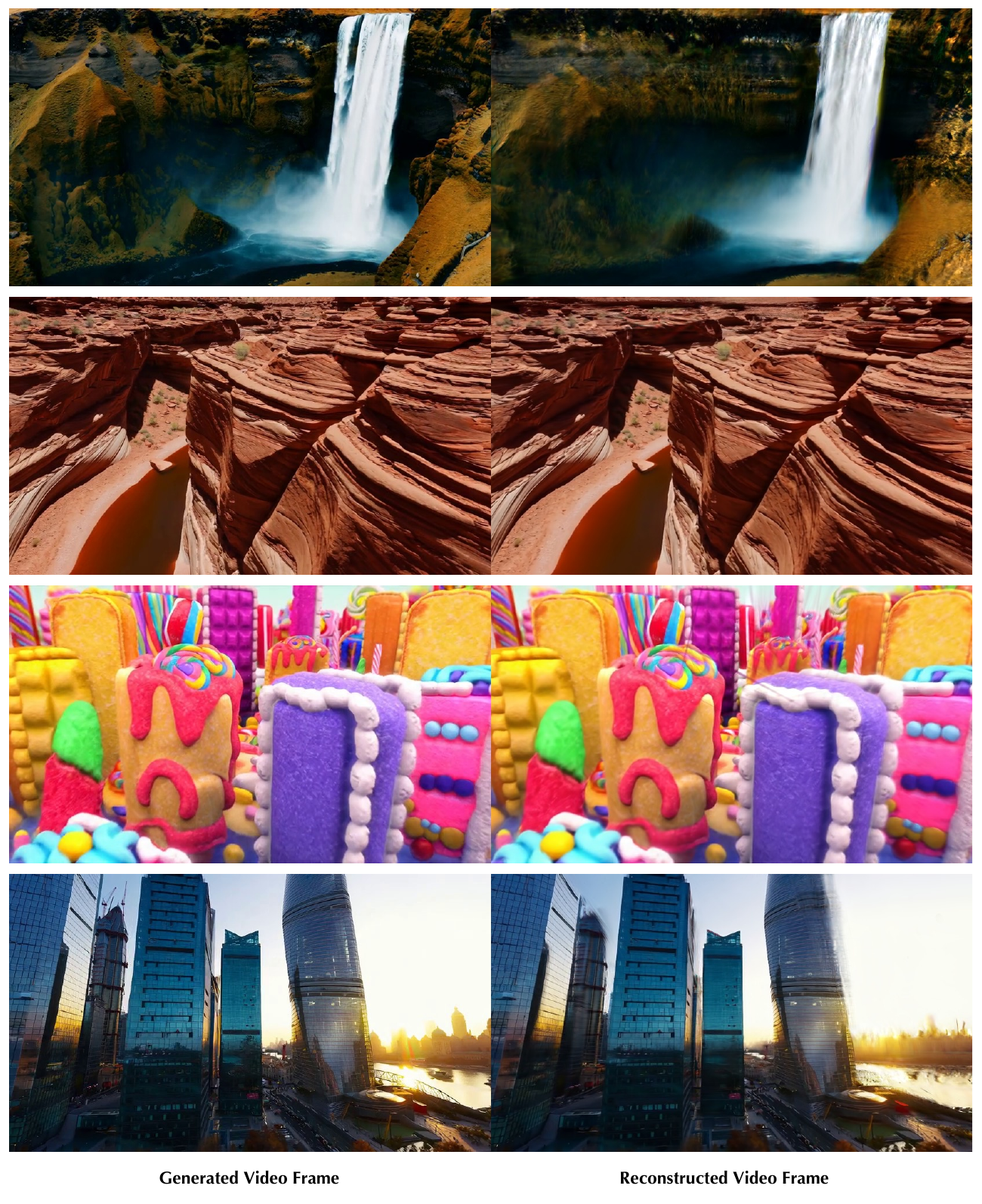}
  \caption{\textbf{High-Fidelity Reconstruction from World-R1.} The strict 3D consistency preserved by \method enables the recovery of a dense and accurate 3D scene representation, free from the floaters and distortion observed in baseline methods.}
  \label{fig:3dgs_success} 
\end{figure}

\begin{figure}[t]
  \centering
  \includegraphics[width=\textwidth]{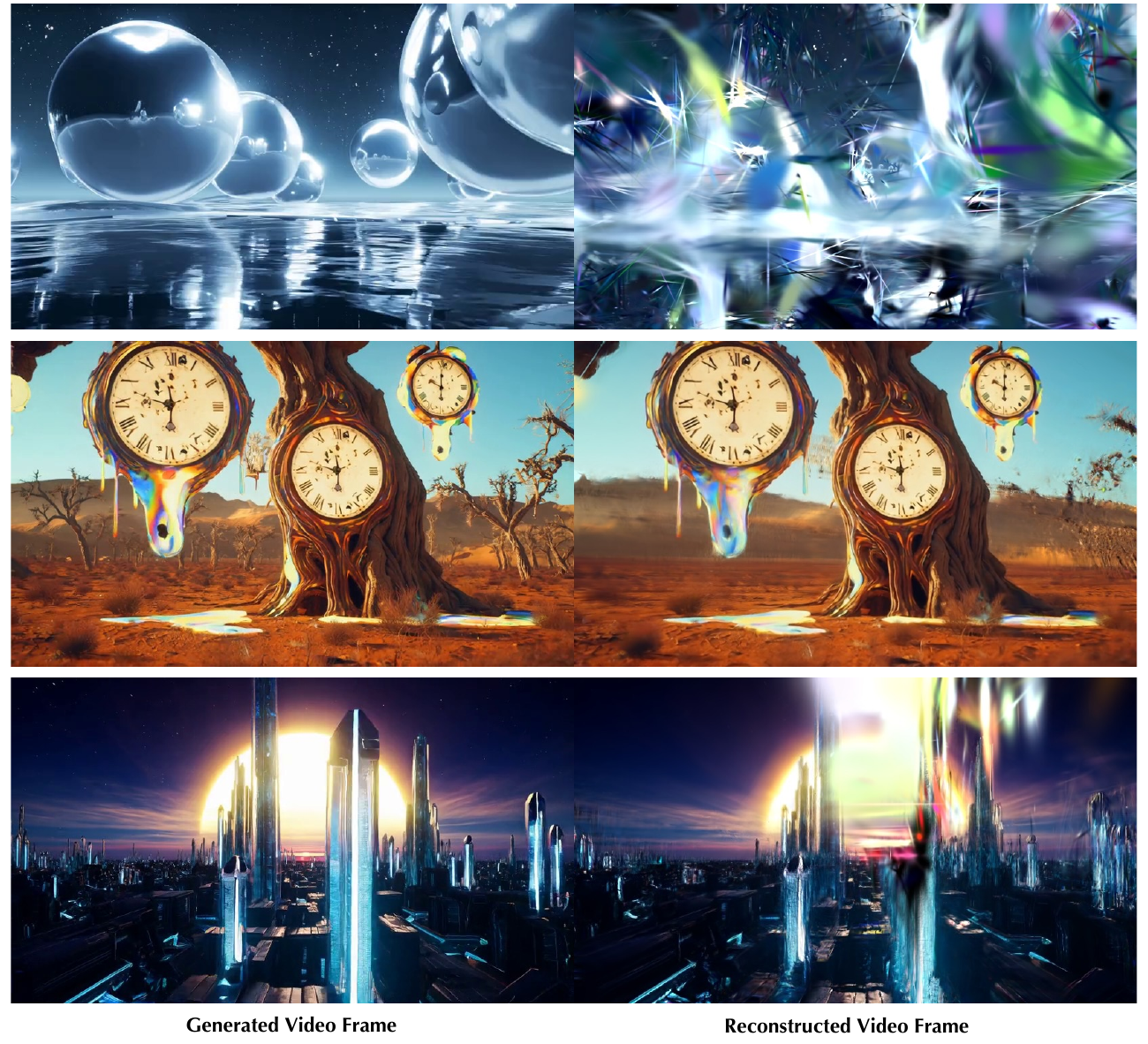}
  \caption{\textbf{Reconstruction Failure due to Geometric Inconsistency.} Visualizations of some reconstruction results derived from the videos of Wan 2.1~\cite{wan2025wan}. Due to temporal warping and object morphing in the source video, the 3D optimization fails to form a coherent structure, resulting in significant noise, sparse point clouds, and visual artifacts.}
  \label{fig:3dgs_failure} 
\end{figure}

\noindent\textbf{Dynamic video results.}
A core challenge in aligning video generation models with 3D constraints is the potential suppression of non-rigid dynamics.
However, thanks to our periodic decoupled training strategy and the utilization of the dynamic data subset, \method successfully disentangles camera motion from object motion. As shown in \cref{fig:dynamic_qualitative}, our model retains the high-entropy generative capabilities of the foundation model.
These results confirm that \method as a robust world simulator, enforcing geometric consistency on the static environment while permitting correct physical dynamics for non-rigid objects.

\noindent\textbf{3D consistency via 3D reconstruction.}
To explicitly validate the geometric integrity of the synthesized worlds, we conduct a reconstruction-based analysis. We compare the original videos generated by \textit{World-R1} with their corresponding re-renderings derived from 3DGS representation (as detailed in the reward formulation in \cref{sec:reward_details}).
Since 3DGS optimization assumes a static scene observed from moving viewpoints, it is highly sensitive to geometric inconsistencies. 
As illustrated in \cref{fig:3dgs_failure} and \cref{fig:3dgs_success}, the difference in reconstruction quality serves as a direct proxy for the underlying video consistency.
The high visual fidelity of the 3DGS re-renderings closely matching the original generated content, demonstrates that \method produces 3D consistent video sequences that adhere to 3D constraints, effectively minimizing geometric hallucinations.

\end{document}